\documentclass[journal]{IEEEtran}
\ifCLASSINFOpdf
\else
\fi

\usepackage{mathrsfs}
\usepackage{color}
\usepackage{amsfonts,amssymb}
\usepackage{amsfonts}
\usepackage{subfigure}
\usepackage{booktabs}
\usepackage{graphicx}
\usepackage{amsmath}
\usepackage{enumerate}
\usepackage{algorithm}
\usepackage{algorithmic}
\usepackage{bm}
\usepackage{amsthm}
\usepackage{cite} 
\usepackage{multirow,makecell}
\usepackage{tabularx}
\usepackage{threeparttable}
\usepackage{siunitx}

\newtheorem{remark}{Remark}
\newtheorem{theorem}{Theorem}

\newtheorem{assumption}{Assumption}

\newtheorem{definition}{Definition}

\newcommand{\degree}{^\circ}

\begin{document}

\title{
Intelligent Physical Attack Against Mobile Robots With Obstacle-Avoidance}
\author{Yushan Li,~\IEEEmembership{Student Member,~IEEE}, Jianping He,~\IEEEmembership{Senior Member,~IEEE}, Cailian Chen,~\IEEEmembership{Member,~IEEE}\\ and Xinping Guan,~\IEEEmembership{Fellow,~IEEE}
	\thanks{
	The authors are with the Department of Automation, Shanghai Jiao Tong University, Shanghai 200240, China; Key Laboratory of System Control and Information Processing, Ministry of Education of China, Shanghai 200240, China; Shanghai Engineering Research Center of Intelligent Control and Management, Shanghai 200240, China. E-mail address: \{yushan\_li, jphe, cailianchen, xpguan\}@sjtu.edu.cn. 
	}
}


\maketitle

\begin{abstract}
The security issue of mobile robots has attracted considerable attention in recent years. 
In this paper, we propose an intelligent physical attack to trap mobile robots into a preset position by learning the obstacle-avoidance mechanism from external observation. 
The salient novelty of our work lies in revealing the possibility that physical-based attacks with intelligent and advanced design can present real threats, while without prior knowledge of the system dynamics or access to the internal system. 
This kind of attack cannot be handled by countermeasures in traditional cyberspace security. 
To practice, the cornerstone of the proposed attack is to actively explore the complex interaction characteristic of the victim robot with the environment, and learn the obstacle-avoidance knowledge exhibited in the limited observations of its behaviors. 
Then, we propose shortest-path and hands-off attack algorithms to find efficient attack paths from the tremendous motion space, achieving the driving-to-trap goal with low costs in terms of path length and activity period, respectively. 
The convergence of the algorithms is proved and the attack performance bounds are further derived. 
Extensive simulations and real-life experiments illustrate the effectiveness of the proposed attack, beckoning future investigation for the new physical threats and defense on robotic systems. 
\end{abstract}

\begin{IEEEkeywords}
Intelligent attack, mobile robots, intentional learning, obstacle-avoidance.
\end{IEEEkeywords}

\IEEEpeerreviewmaketitle

\section{Introduction}
With integrated communication, computation and control to support the operations in the physical world, the mobile robots can be seen as a typical cyber-physical system (CPS), attracting extensive attention thanks to their excellent flexibility and scalability. 
From UAVs to UGVs, either single or multiple coordinated, mobile robots are becoming more and more pervasive in industrial and military fields, e.g., logistics transportation, environment exploration, surveillance and reconnaissance. 
With this ever-increasing popularity, their security becomes critical yet imperatively challenging \cite{choi2018detecting}, given numerous possible malicious or unanticipated behaviors.

This paper investigates a novel intelligent physical attack against mobile robots without relying on any prior knowledge. 
The ultimate goal of the attacker is to learn the obstacle-avoidance mechanism of a mobile robot from external observation, and then leverage it to fool the target robot into a preset trap. 
This problem is inspired by the intrinsic interaction nature of a robot with the physical environment and its security issues. 
On the one hand, prior works concerning the security of mobile robots are mainly developed from the perspective of cyberspace \cite{ding2018survey,sanchez2019bibliographical}, while physical attacks are less noticed yet also critically important. 
It is universally needed for mobile robots in the real world to interact with the physical environment \cite{pandey2017mobile}. 
By disguising as an obstacle and physically approaching a robot, the resulting obstacle-avoidance behavior is inevitable, providing support to trap the robot. 
On the other hand, numerous defense techniques assume possible attacks are powerful beforehand \cite{pasqualetti2012consensus,pasqualetti2013attack}, obtaining robust security performance while incurring more costs, e.g., hardware computation burdens. 
Indeed, if an attack is figured to be almost impossible to launch, then there is no need to design sophisticated countermeasures along with performance degradations, better guiding the systems defense design.

Despite some similar features between the attack problem in this paper and classic pursuit-evasion \cite{vidal2002probabilistic,kolling2009pursuit,mejia2019solutions} and herding problems \cite{lee2017autonomous,varava2017herding,licitra2019single}, we point out there are still major differences in their application scenarios and basic model formulation. 
First, the physical attack here is not purely about controlling a robot to capture or track the victim robot, but making the victim robot move to a preset deterministic position. 
Second, unlike the aforementioned works that usually assume complete or partial knowledge about interaction models is available, the attacker here has no prior knowledge about the victim robot.
It needs to actively learn the necessary obstacle-avoidance knowledge first, making the attack more generalizable and applicable to a majority of obstacle-avoidance mechanisms, which is a significant and unique component of this paper. 
Besides, compared with directly intercepting (e.g., missile guidance \cite{ratnoo2011lineofsight}) or decoding-and-manipulating the communication messages (especially when the control-communication is protected with strong encryption \cite{darup2018encrypted,alexandru2019encrypted}), the proposed physical attack is more feasible, stealthy and economically viable, presenting serious threats. 

To systematically design and evaluate the intelligent physical attacks on mobile robots, 
three major challenges must be addressed. 
First, the attacker is unaware of the prerequisite information that supports the attack (like the obstacle detection range and goal position), making it an impediment to formulating the attack model by traditional techniques, e.g., parameter identification and model predictive control. 
Second, the attacker and victim robots are mutually influenced in the motion space, incurring difficulties to find the feasible solution space that meets the attack requirements. 
Third, even if a feasible solution space is found, it is hard to optimize strategy designs to achieve the final purpose with low attack costs. 

This paper is an extension of the preliminary work presented in \cite{lys}, providing a detailed and rigorous treatment of model learning, attack design and performances analysis, as well as significant novel simulation results. 
The main contributions of this paper are as follows:

\begin{itemize}
\item 
We investigate the possibility of achieving an intelligent and advanced physical attack against mobile robots, 
merely utilizing external observations and not relying on any prior information of the system dynamics. 
Taking the universal obstacle-avoidance mechanism as the attack interface, we provide a systematic attack scheme including knowledge-learning and attack implementation. 
\item We propose an intentional excitation based learning approach to acquire the obstacle-avoidance knowledge. 
By characterizing the obstacle-avoidance behaviors and disguising the attacker as an obstacle, we demonstrate how to excite various avoidance reactions of the victim robot. 
Based on observations over this process, we establish the featured data pairs that reflect the underlying mechanism and further regress it by learning-based methods. 
\item We design two driving-to-trap attack algorithms by taking the attack path length and activity period as the objectives, respectively. 
We prove the convergence of the attack algorithms from the perspective of the descent search. 
The performance bounds of the algorithms are further derived concerning the optimal cost in theory. 
Extensive simulations and real-life experiments are conducted to illustrate the effectiveness of the proposed attack. 
\end{itemize}

In a larger sense, this paper reveals the vulnerability of system security from external observation and excitation. 
As the obstacle-avoidance behavior is indispensable for a mobile robot to interact with the environment, the proposed attack is hard to defend when the robot has no prior knowledge about the malicious behavior, where the attacker is not always consistently but intermittently active during the process. 
Therefore, our work lays the foundation to explore more potential and advanced attacks against mobile robots from physical aspects, and provides deeper insight to guide the defense design in the future.

\begin{figure}[t]
\centering
\includegraphics[width=0.47\textwidth]{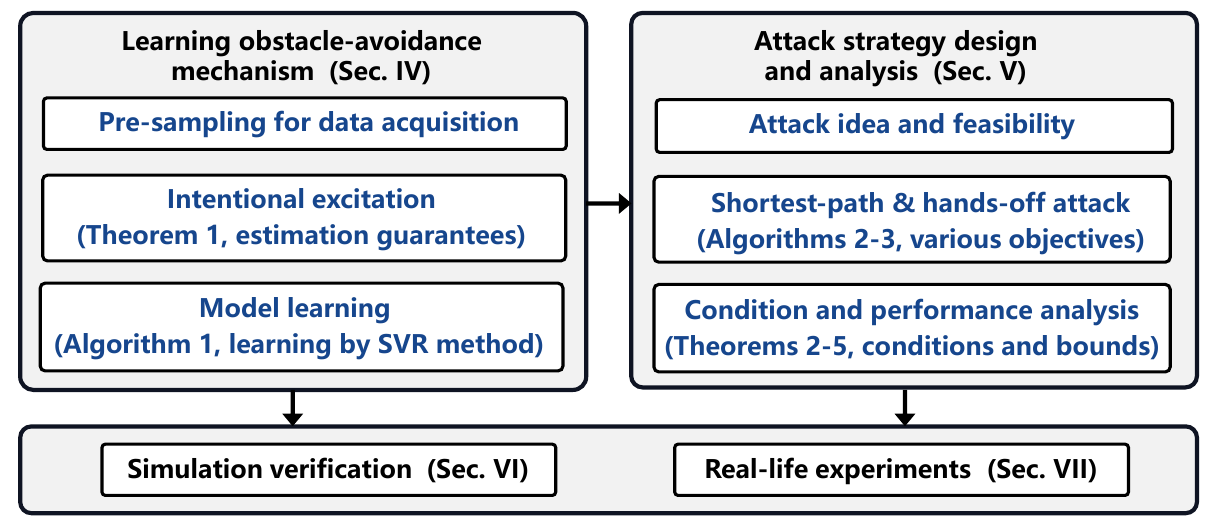}
\caption{Roadmap of the main results in this paper. }
\label{fig:roadmap}
\vspace{-10pt}
\end{figure}

The rest of this paper is organized as follows. 
To begin with, Section \ref{r-work} presents an overview of the relevant literature. 
In Section \ref{sec:pre}, the kinematics and obstacle-avoidance modeling for both holonomic and non-holonomic mobile robots are introduced. 
The learning scheme for the obstacle-avoidance mechanism is proposed in Section \ref{learning-scheme}. 
Section \ref{attack-strategy} presents the attack strategies with performance analysis. 
Simulations and experiments are shown in Section \ref{sec-simulation} and Section \ref{sec:experiment}, respectively. 
Finally, Section \ref{sec-conclusion} concludes this paper. 
A roadmap of the main results of this paper is shown in Fig.~\ref{fig:roadmap}.

\section{Related Work}\label{r-work}
There are mainly two related topics: CPS security and obstacle-avoidance mechanism. 
Given the richness of both fields, a brief overview of some representative works is given. 

\textbf{CPS security.} 
The security shows the ability of a system to govern malicious behaviors or unanticipated events \cite{zhu2015game}. 
In the literature, attacks against mobile robots are mainly from cyberspace, and they can be roughly divided into three categories: denial of service (DoS), replay, and deception attacks \cite{ding2018survey}. 
In these attacks, communication channels are jammed/disrupted \cite{feng2019secure}, or the control data and measurements are compromised/altered \cite{su2018supervisor}, and thus compromising the systems in critical and adversarial scenarios. 
Many efforts have been devoted to designing corresponding countermeasures. 
For instance, \cite{pasqualetti2013attack} characterized the undetectable and unidentifiable attacks and designed corresponding detection filters. 
\cite{chang2018secure} considered the secure estimation problem where the set of attacked nodes can change with time. 
The motion synchronization methods were developed for robot formation considering the presence of robot malfunctions \cite{liu2017self,liu2018failure} or undesirable formation variation \cite{dong2014time,dong2017time}. A group of robots were deployed to herd an adversarial aerial attacker to defend safety-critical infrastructure in \cite{chipade2019herding}. 
However, most existing research relied on the assumption that the attacker has some knowledge or can access the robot system. 
For example, malicious agents have knowledge of the system structure or nodes' states \cite{pasqualetti2012consensus}, or the packets transmitted over the network are corrupted \cite{silvestre2017stochastic,dong2019false}. 
These assumptions may no longer hold given the cryptographic solutions in practice. 

In recent years, stealthy physical attacks emerged and presented real threats, where the physical components were targeted to compromise or even destroy the system. 
For instance, the GPS sensor readings can be disturbed by GPS spoofing attack \cite{tippenhauer2011requirements,bianchin2019secure}. 
Designed acoustic noises can alter gyroscopic sensor data, leading to drone crashes \cite{son2015rocking}. 
Even the important values stored in memory (e.g., EEPROM, Flash memory) can be corrupted by heating a memory cell while the device is without any damage \cite{skorobogatov2009local}. 
Compared with cyber attacks, these physical attacks have a wider range of applicability, while the traditional detection techniques from the computer security community are usually not effective in handling them \cite{choi2018detecting}. 
Note that these emerging physical attacks are still elementary and not sufficiently intelligent. 
For instance, the attacks were designed against a specific transducer by utilizing its sensing mechanism, and cannot be generalized. 
In addition, these physical attacks aiming to disturb the system performance were designed with an ``open-loop'' flavor and without sophisticated control design, which motivates this work to investigate possible physical attacks with intelligence. 

\textbf{Obstacle-avoidance mechanism.} 
Obstacle-avoidance is the major interface for mobile robots to interact with the outside environment. 
Numerous methods have been developed in the literature, for example, potential field based approach \cite{khatib1986real}, genetic algorithm based approach \cite{merchan2006fuzzy}, fuzzy logic based approach \cite{li2013design}, neural network based approach \cite{li2009neural}, etc (see \cite{goerzen2010survey,mohanan2018survey} for a detailed review). 
According to their algorithm characteristic, similar to \cite{tam2009review}, we divide these algorithms into two main types: 
deterministic (e.g., artificial potential method, learning-based method) and search-based (e.g., dynamic window approach, genetic approach, evolutionary algorithm). 
The former is determined-model driven\footnote{Although learning-based method is commonly said data-driven, the model is generally injective mapping once the training is completed.} and the solution for the current situation is unique and found quickly. 
The key idea of the latter one is to search for all feasible solutions and select the best one according to an evaluation function, which is time-consuming. 
No matter what obstacle-avoidance method is used, the essence of circumventing the obstacle with certain trajectory adjustment towards its goal is identical and directly leaks to external observation. 
This nature is captured and leveraged in our attack design.

In summary, it remains an open yet critical issue to investigate what kind of intelligent and generalized attack is available from the physical side, while without prior knowledge about the system.

\begin{table}[t] 
 \caption{\label{tab:test}Notation Definitions} 
 \begin{tabular}{ll}
  \toprule 
  Symbol  & Definition  \\ 
  \midrule
  $\text{R}_\text{a}$ & the abbreviated form of the attack robot\\ 
  $\text{R}_\text{v}$ & the abbreviated form of the victim robot\\
  $f$ & the obstacle-avoidance mechanism\\ 
  $\mathcal{J}$ & the observed dataset for learning the knowledge about $f$\\ 
  $p_{\vartheta}$ & \makecell[t{p{6.8cm}}]{the position variable. $\vartheta=a,v,g,t,e$ represent the position of $\text{R}_\text{a}$, $\text{R}_\text{v}$, the task goal, the preset trap, and the attack entry point, respectively} \\
  $\mathcal{T}_{\vartheta}$ & \makecell[t{p{6.8cm}}]{the trajectory variable. $\vartheta\!=\!g,a,m$ represent the ideal trajectory of $\text{R}_\text{v}$ to $p_{g}$ without being attacked, the ideal attack trajectory from $p_e$ to $p_{t}$, and the instantaneous motion direction trajectory of a robot, respectively}\\
  $\mathcal{A}_{\vartheta}$ & \makecell[t{p{6.8cm}}]{the area variable, representing a set of specific positions } \\ 
  $(D,\bm{\alpha})$ & \makecell[t{p{6.8cm}}]{the obstacle-avoidance detection range, where $D$ is the radius and $\bm{\alpha}$ is the angle range} \\  
  $d(p_1,p_2)$ & the distance between $p_1$ and $p_2$, i.e., $\| {p_2 - p_1} \|_2$\\
  $d_T(\mathcal{T}_{\vartheta},p)$ & the minimal distance from a position $p$ to $\mathcal{T}_{\vartheta}$\\
  $I_T(\mathcal{T}_{\vartheta},p)$ & the indicative function to judge whether $p$ is on $\mathcal{T}_{\vartheta}$ \\  
  \bottomrule 
 \end{tabular} 
\end{table}

\section{Preliminary and Problem Formulation}\label{sec:pre}
In this section, we first introduce some basics about the kinematics of a mobile robot. The motion dynamics of both holonomic and non-holonomic mobile robots are presented. 
Following this, we show what an important role the obstacle-avoidance plays. 
At last, our problem of interest is formulated. 
Some important notation and definitions are given in Table \ref{tab:test}. 

\subsection{Motion Control for Mobile Robots}
It should be mentioned that there are generally two kinds of mobile robots: non-holonomic (car-like wheeled mobile agents, unicycles, etc.) and holonomic. 
Although the former one appears more common in daily life, their instantaneous movement is restricted laterally \cite{listmann2009consensus}, making the control more challenging than that of the latter kind. 

For a non-holonomic mobile robot, the posture is usually represented by its position $(x,y)$ and orientation $\theta \in [0, 2\pi)$ with respect to X axis, denoted as $p=[x, y, \theta]^\mathsf{T}$. 
The motion is controlled directly by linear velocity $v$ and angular velocity $\omega$ or velocities of two driving wheels, which are equivalent to each other. 
Then, its kinematics is modeled as
\begin{equation}\label{old}
\left\{ 
\begin{aligned}
\dot x &= v\cos(\theta),\\
\dot y &= v\sin(\theta),\\
\dot \theta&=  \omega.
\end{aligned}
\right.
\end{equation}
Accordingly, the discrete form of (\ref{old}) is given by 
\begin{equation}\label{E001}
\left\{ 
\begin{aligned}
x(t_i^c) &= x(t_{i - 1}^c) + v(t_{i - 1}^c) \cdot \cos (\theta (t_{i - 1}^c))\cdot T,\\
y(t_i^c) &= y(t_{i - 1}^c) + v(t_{i - 1}^c) \cdot \sin (\theta (t_{i - 1}^c))\cdot T,\\
\theta (t_i^c) &= \theta (t_{i - 1}^c) + \omega (t_{i - 1}^c) \cdot T,
\end{aligned}
\right.
\end{equation}
where  $T$ is the motion control period, 
and $t^c_i$ ($i=1,2,\cdots$) denotes the control time instant satisfying $t^c_{i}-t^c_{i-1}=T$. 
The motion control aims to drive the robot toward its desired posture $p^*=[x^*, y^*, \theta^*]^\mathsf{T}$ to achieve predefined tasks by designing the velocities $v$ and $\omega$. 
Define the instantaneous posture error as $p_e=p^*-p =[x_e, y_e, \theta_e]^\mathsf{T}$, and the controller can be represented as 
\begin{equation}\label{eq:motion-controller-1}
\left[ {\begin{array}{*{20}{c}}
v\\
\omega
\end{array}} \right]= \left[ {\begin{array}{*{20}{c}}
u_v(p_e,\dot{p}_e)\\
u_{\omega}(p_e,\dot{p}_e)
\end{array}} \right],
\end{equation}
where $u_v$ and $u_{\omega}$ are designed to control $v$ and $\omega$ such that the tracking error $\|p_e\|_2$ goes to zero. 
Note that (\ref{eq:motion-controller-1}) represents a general first-order motion control for nonholonomic mobile robots. 
Many works have been developed with stability guarantees \cite{samson1993time,astolfi1999exponential,chen2010leader}. 
They usually require the measurements of $\dot{p}^*$ and are of proportional-derivative-trigonometric type\footnote{The introduction of the trigonometric function is mainly caused by the decomposition of linear velocity. 
For example, in \cite{chen2010leader}, the motion controller is designed as $v=v^* \cos{\theta^e}+{k_x} x^e$, $\omega={\omega^*} + k_{\theta}\theta^e + v^* y^e \sin{\theta^e}/\theta^e$.}. 

Considering the nonlinear characteristic of non-holonomic robots, it is common to introduce a ``hand'' position ${\bm{h}}\!=\![{x_h,y_h}]^\mathsf{T}\!=\![{x\!+\!d_h\cos\theta,y\!+\!d_h \sin\theta}]^\mathsf{T}$ to simplify the motion control,  
where $d_h$ is a distance starting from $p$ along the orientation axis. 
Taking ${\bm{h}}$ into (\ref{old}), one has 
\begin{equation}\label{E14}
 \left[ {\begin{array}{*{20}{c}}
{{{\dot x}_h}}\\
{{{\dot y}_h}}
\end{array}} \right] = \left[ {\begin{array}{*{20}{c}}
{\cos \theta }&{ -d_h\sin \theta }\\
{\sin \theta }&{d_h\cos \theta }
\end{array}} \right]\left[ {\begin{array}{*{20}{c}}
v\\
\omega 
\end{array}} \right].
\end{equation}
By output feedback linearization, the controller for $(v,\omega)$ can be designed as \cite{lawton2003decentralized}
    \begin{equation}\label{eq:linearization}
      \left[ {\begin{array}{*{20}{c}}
      {{v}}\\
      {{\omega }}
      \end{array}} \right] = \left[ {\begin{array}{*{20}{c}}
      {\cos {\theta}}&{\sin {\theta}}\\
      { - \frac{1}{{{d_h}}}\sin {\theta}}&{\frac{1}{{{d_h}}}\cos {\theta }}
      \end{array}} \right]\left[ {\begin{array}{*{20}{c}}
      {{u_{x}(x_e,\dot{x}_e)}}\\
      {{u_{y}(y_e,\dot{y}_e)}}
      \end{array}} \right],
    \end{equation}
which gives the following simplified point regulation of the hand position
    \begin{equation}\label{eq:h-first-order}
      \left[ {\begin{array}{*{20}{c}}
      \dot{x}_h\\
      \dot{y}_h
      \end{array}} \right]= \left[ {\begin{array}{*{20}{c}}
      u_x(x_e,\dot{x}_e)\\
      u_{y}(y_e,\dot{y}_e)
      \end{array}} \right],
    \end{equation}
The controller (\ref{eq:h-first-order}) is decoupled and designed in proportional-derivative form to guarantee the stability of $\bm{h}$. 

There are a few works focusing on directly controlling the applied torque and force for the wheels based on $p_e$ \cite{lawton2003decentralized,hernandez2020consensus}, which is essentially a second-order controller. 
In this paper, we do not focus on specific motion controller design, but emphasize that the desired posture of the robot can be affected to achieve our attack no matter what motion controller the robot uses. 



As for a holonomic mobile robot, the kinematics in two directions is independent and its discrete form is given by 
\begin{equation}\label{E002}
\left\{ 
\begin{aligned}
x(t_i^c) &= x(t_{i - 1}^c) + v_x(t_{i - 1}^c) \cdot T,\\
y(t_i^c) &= y(t_{i - 1}^c) + v_y(t_{i - 1}^c) \cdot T,
\end{aligned}
\right.
\end{equation}
where $v_x$ and $v_y$ are velocities along X and Y axis directions, respectively. 
Since the motion of a holonomic robot is a direct composition of the motions in two directions, the orientation is usually neglected. 
Therefore, different from (\ref{eq:motion-controller-1}), here the motion control is directly formulated as
\begin{equation}\label{eq:motion-controller-2}
\left[ {\begin{array}{*{20}{c}}
v_x\\
v_y
\end{array}} \right]= \left[ {\begin{array}{*{20}{c}}
u_x(p_x,\dot{p}_x) \\
u_y(p_y,\dot{p}_y)
\end{array}} \right].
\end{equation}

The proposed intelligent attack in this paper applies to both kinds of robots. 
Hereafter, we mainly illustrate our work on non-holonomic robots. 
It is easy to extend the results to holonomic ones due to the simple motion characteristic, and they will be briefly discussed. 

\subsection{Obstacle-avoidance Behavior Modeling}
The obstacle-avoidance mechanism is the major interface for a robot to interact with the physical environment, and is universally needed for the robot to work. 
Let $p_{ob}$ and $v_{ob}$ be the state and velocity of the obstacle, respectively. 
Considering that avoidance behavior is mainly determined by the relative posture and motion between the robot and the obstacle, we formulate it as the following general function
\begin{equation}\label{local}
u_{ob}= f({p_{ob}}-{p},p^{*}-p,v_{ob},{v}),
\end{equation}
where $p^{*}$ is the instantaneous desired posture determined by a specific task (e.g., it can represent the target leader posture in formation control). 
In some algorithm designs, the influence of $(p^{*} - p)$ is negligible when $p_{ob}$ is very close to $p$. 

Next, we characterize the obstacle-avoidance behavior under (\ref{local}). 
Let $\mathcal{A}_d(p)$ be the obstacle detection area of a robot. 
The obvious characteristic of $f$ is its boundedness regardless of the detailed design, which can be described as 
\begin{equation}
\left\{ {\begin{aligned}
&\|f( \cdot )\|_2 = 0,&&~\text{if}~ p_{ob}\notin \mathcal{A}_d(p),\\
&{0 \le \|f( \cdot ) \|_2 \le \bar{u}_{ob}},&&~\text{if}~ p_{ob}\in \mathcal{A}_d(p),\\
\end{aligned}} \right.
\end{equation}
where $\bar{u}_{ob}$ is the output bound. 
Another salient characteristic is the directionality of the obstacle-avoidance behavior. 
For ease of expressions, let $\mathcal{T}_{\vartheta}$ be a general notation for a simple curve (trajectory) that divides the X-Y plane into two parts, $\mathcal{T}_{\vartheta}^{+}$ and $\mathcal{T}_{\vartheta}^{-}$ (the subscript $\vartheta$ will be replaced when indicating a specific trajectory). 
Define $I_T(\mathcal{T}_{\vartheta},p)$ as the indication function
\begin{equation}\label{eq:indicative}
{I_T}(\mathcal{T}_{\vartheta},p) = \left\{ 
\begin{aligned}
&0,&&\text{if}~p\in \mathcal{T}_{\vartheta};\\
&+1,&&\text{if}~p \in \mathcal{T}_{\vartheta}^{+};\\
&-1,&&\text{if}~p \in \mathcal{T}_{\vartheta}^{-}.
\end{aligned} \right.
\end{equation}
Then, when an obstacle occurs in the detection range at moment $k$ (i.e., $p_{ob}\in\mathcal{A}_d(p(k))$) and $\mathcal{T}_m(p)$ is the instantaneous tangent line of the robot's motion, the obstacle-avoidance behavior satisfies 
\begin{equation}\label{eq:cond-2}
I_T(\mathcal{T}_m(p(k)),p_{ob})\!\cdot\! I_T(\mathcal{T}_m(p(k)),p(k+1))<0,
\end{equation}
which indicates that the robot will move towards the opposite side of the obstacle taking $\mathcal{T}_m(p(k))$ as a boundary.

There are some notable points about the obstacle-avoidance modeling in some complicated cases. 
\begin{itemize}
\item Geometric shape. 
In practice, the obstacle is detected by the sensors as a series of dense points, and the avoidance reaction is based on processing these points by certain computation rules. 
For example, they can be enclosed by a convex geometric shape that has a simple representation (e.g., a circle) and the robot moves around the central point with a known radius \cite{zhu2020velocity}. 
\item Multiple obstacles. 
When multiple obstacles are involved, the robot will evaluate the influences of the obstacles and obtain their composite impact. 
In some simple cases, the robot can directly take the nearest obstacle into its avoidance procedures at the moment, which always guarantees a safe avoidance behavior. 
\end{itemize}


The avoidance model (\ref{local}) covers a majority of algorithms that are based on local detection range and use the obstacle's instantaneous motion state as the input. 
Considering when the obstacle is moving, some advanced methods are developed to improve the avoidance efficiency by predicting and evaluating the motion of the obstacle \cite{jaillet2004prm,large2005navigation,shim2007evasive,du2011robot}. 
Nevertheless, these methods usually require some prior knowledge, e.g., the obstacles are assumed to move along deterministic paths \cite{shim2007evasive} or satisfy certain nominal kinematic constraints \cite{du2011robot}. 
Based on the above analysis, we make the following assumption throughout this paper. 
\begin{assumption}\label{assum:ob}
The obstacle-avoidance detection range of the victim robot is local, and the obstacle-avoidance algorithm it uses can be modeled by (\ref{local}) and is an injective mapping. 
Besides, the observations of the attack robot on the victim robot are corrupted with Gaussian noise $\mathcal{N}(0,\sigma^2)$. 
\end{assumption}


\begin{remark}
For the class of moving obstacle avoidance algorithms with prior knowledge acquired, 
the attack idea and procedure proposed in this paper can still be applied with additional computation costs and higher algorithm complexity, which come from the two aspects. 
First, the historical positions and velocities of the obstacle itself need to be considered for modeling and learning the mechanism $f$. 
Second, the search space during attack implementation should also be increased according to the time horizon used for learning $f$. 
\end{remark}

\subsection{Problem of Interest}
Denote the malicious attack robot and the victim robot as $\text{R}_\text{a}$ and $\text{R}_\text{v}$, respectively, and let $f$ represent the obstacle-avoidance mechanism. 
Our work applies to a basic and representative application scenario where $\text{R}_\text{v}$ is performing a go-to-goal task and able to detect and avoid obstacles occurring in its surroundings by $f$. 
Concretely, $\text{R}_\text{a}$ aims to fool $\text{R}_\text{v}$ into a preset trap $p_t$ by disguising as an obstacle, while it can observe (or sense) $\text{R}_\text{v}$ and has limited moving capability in this process. 
Define $\mathcal{J}$ as the observed dataset reflecting $f$, and the attack is mathematically transformed to solve the following problems. 
\begin{itemize}
\item Learning $f$: given $\mathcal{J}$, find the mapping relation $\phi$ that embodies $f$ by solving 
\begin{equation}\label{f-problem}
\mathop {\min }\limits_{\phi (\mathcal{J}) \mapsto f} {\| {f - \hat f} \|_2 }. 
\end{equation}
\item Attack strategy design: given $\hat f$, find a group of attack inputs $\bm{u}_{a,0:H}$ that drives $\text{R}_\text{v}$ into $p_t$ by solving 
\begin{equation}\label{attack-problem}
\begin{aligned}
\textbf{P}_\textbf{0}:~~~~
\mathop {\min }\limits_{H,{\bm{u}_{a,0:H}}}~&C_{\vartheta}(\bm{u}_{a,0:H}) \\
\text{s.t.}~~~~&{\| {{p_v}(H) - p_t} \|_2} \le \delta,\\
\end{aligned}
\end{equation}
where $C_{\vartheta}$ is the attack cost (e.g., attack path length or activity period) and $\delta\ge0$ is a constant that indicates attack termination. 
\end{itemize}

\begin{figure}[t]
\centering
\includegraphics[width=0.5\textwidth]{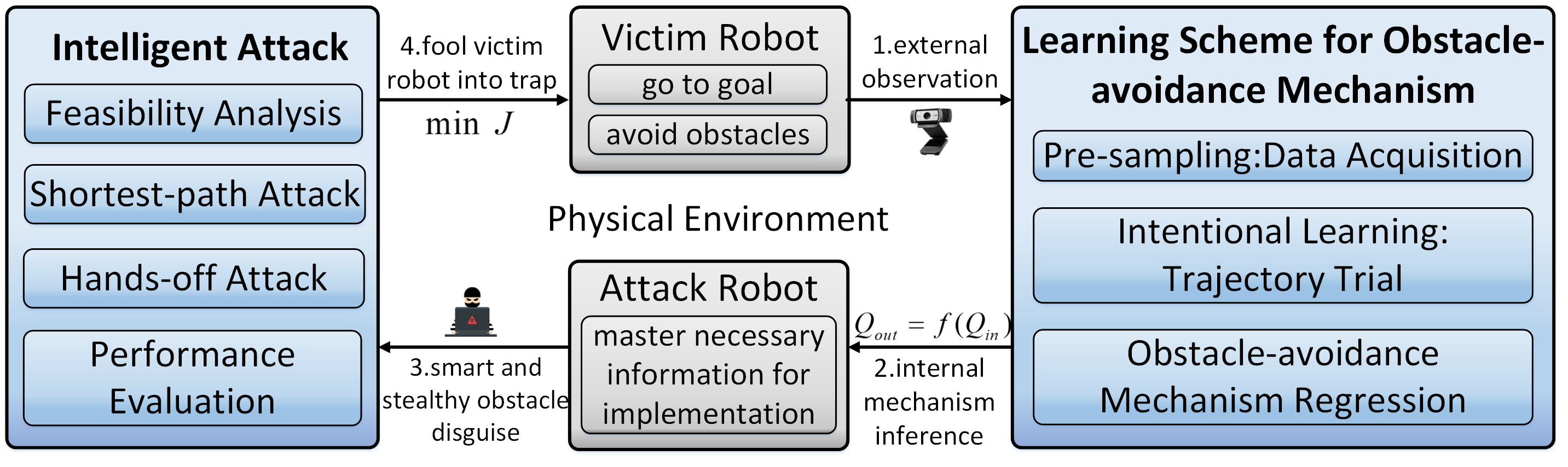}
\caption{The architecture of the proposed intelligent physical attack. The primary goal of $\text{R}_\text{a}$ is to learn the knowledge about the $f$. 
Then, with a learned $f$, efficient attack strategies for different attack requirements are designed to achieve the final purpose.} 
\label{framework}
\vspace*{-8pt}
\end{figure}

To solve problem (\ref{f-problem}), the key point is to capture the essential principle about how the mobile robot will adjust its trajectory after an obstacle is detected nearby, and construct a corresponding dataset from observations. 
For problem (\ref{attack-problem}), it is extremely hard to get an optimal analytic solution due to the implicit form of $f$. 
Nevertheless, borrowing ideas from the sampling-based approaches \cite{lan2013planning,hollinger2013sampling}, we are able to design efficient algorithms to obtain suboptimal solutions in terms of different attack costs. 
The whole framework of this paper is shown in Fig.~\ref{framework}.

\section{Learning Scheme for Obstacle-avoidance Mechanism}\label{learning-scheme}
When $\text{R}_\text{a}$ encounters an obstacle within its detection area, it will evaluate the obstacle's influence and take corresponding actions, deviating from its desired trajectory. 
Inspired by this, a learning scheme for the obstacle-avoidance mechanism is proposed. 
This scheme consists of three parts: 
\begin{itemize}
\item Pre-sampling. 
Ideally, the instantaneous motion information of $\text{R}_\text{v}$ (such as orientation, linear, angular and acceleration velocities) can be obtained based on three consecutive position samplings. 
This constitutes the cornerstone of the following steps.

\item Intentional excitation. 
With the ability of observing $\text{R}_\text{v}$'s motion, $\text{R}_\text{a}$ intentionally excites $\text{R}_\text{v}$ by approaching it as a detected obstacle. 
Then, $\text{R}_\text{v}$ is steered to generate various obstacle-avoidance behaviors, enriching the observations to learn the underlying mechanism.  

\item Modeling learning. 
Based on the collected data of the excitation trials, $\text{R}_\text{a}$ computes the featured process variables that exhibit the obstacle-avoidance behavior, and regresses the underlying mechanism (\ref{local}) by learning-based methods (e.g., SVR, RRT). 
\end{itemize}

\subsection{Pre-sampling: Data Acquisition}
Recall that the posture $p$ of a mobile robot is updated every control period $T$, and its trajectory during the period $T$ can be approximated as a straight line. 
Equipped with advanced sensors, $\text{R}_\text{a}$ is able to measure its relative displacement with a moving object. 
Let the sampling period of $\text{R}_\text{a}$ be $\tilde T=NT$, and denote by $t^s_k$ the $k$-th sampling time satisfying $t^s_{k}-t^s_{k-1}=\tilde T$. 
For simplicity, we use the subscript $k$ to denote the sampling time, e.g., $x_k$ represents the $x(t^s_{k})$. 

After three consecutive sampling moments, the instantaneous variables of $\text{R}_\text{v}$'s motion are estimated by
\begin{equation}\label{eq1}
\left\{
\begin{aligned}
{v_{x,k + 1}} &= \frac{x_{k + 1}-x_{k}}{\tilde T},~~~~~~~~\!{v_{y,k + 1}} = \frac{y_{k + 1}-y_{k}}{\tilde T},\\
{v_{k + 1}} &= \sqrt {v_{x,k + 1}^2 + v_{y,k + 1}^2},~{\theta _{k + 1}}=\text{arctan}\frac{v_{y,k + 1}}{v_{x,k + 1}},\\
{a_{k + 1}} &= \frac{{{v_{k + 1}} - {v_k}}}{\tilde T},~~~~~~~~~{\omega _{k + 1}} = \frac{{{\theta _{k + 1}} - {\theta _k}}}{\tilde T},
\end{aligned}
\right.
\end{equation}
where $v, w, a$ represent the linear, angular and accelerated velocities, respectively. 
Note that $\text{R}_\text{a}$ should have the latest three groups of observations to calculate (\ref{eq1}). 
For holonomic mobile robots, this process is much easier with higher precision, and ${v_x}, {v_y}$ can be directly used. 
To make a unified statement, let ${v_1}\!=\!v, {v_2}\!=\!\omega$ if the robot is non-holonomic or ${v_1}\!=\!{v_x}, {v_2}\!=\!{v_y}$ if holonomic. 
\begin{remark}
\label{remark3}
Generally, the motion control period $T$ is very small in practice (e.g., 0.05s or 0.1s), and the sampling period $\tilde T$ is determined by the information sensing and processing ability that $\text{R}_\text{a}$ could afford. 
Note that the influence of observation noises can be largely weakened if one set a larger sampling period and then calculate an average value in a smaller period (e.g., $T$). 
This will not jeopardize the following analysis for the learning and attack procedures. 
\end{remark}

\begin{figure*}[t]
  \centering 
  \setlength{\abovecaptionskip}{0.1cm}
   \subfigure[]{ 
    \label{infer1} 
    \includegraphics[width=0.33\textwidth]{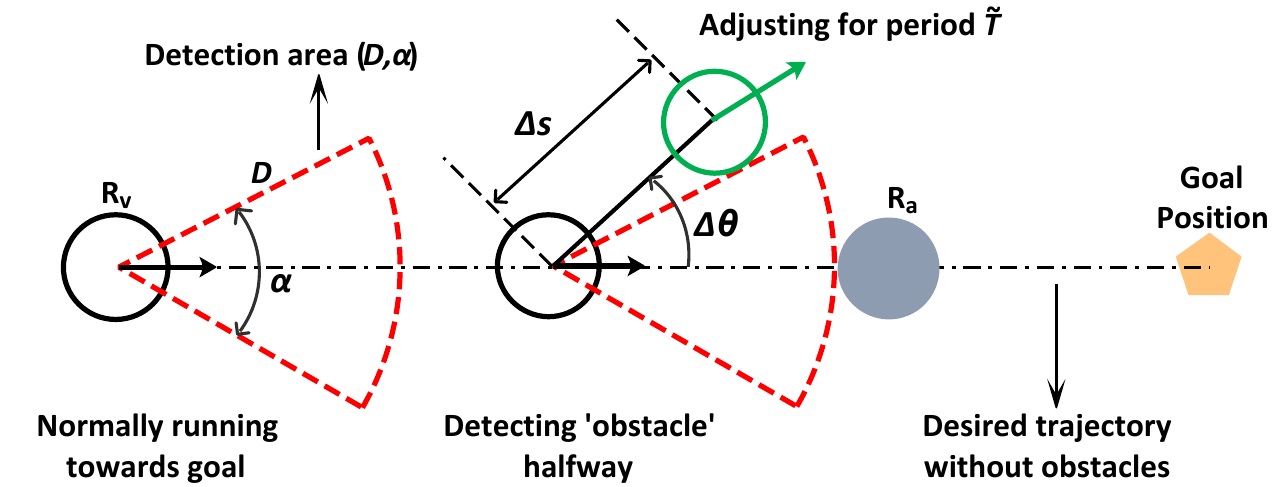}
  } 
  \subfigure[]{ 
    \label{infer2} 
    \includegraphics[width=0.3\textwidth]{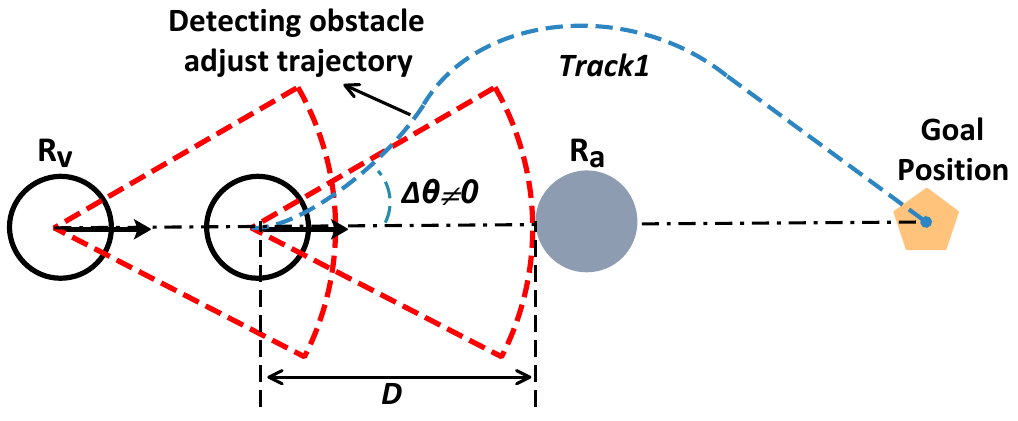} 
  } 
  \subfigure[]{ 
    \label{infer3} 
    \includegraphics[width=0.3\textwidth]{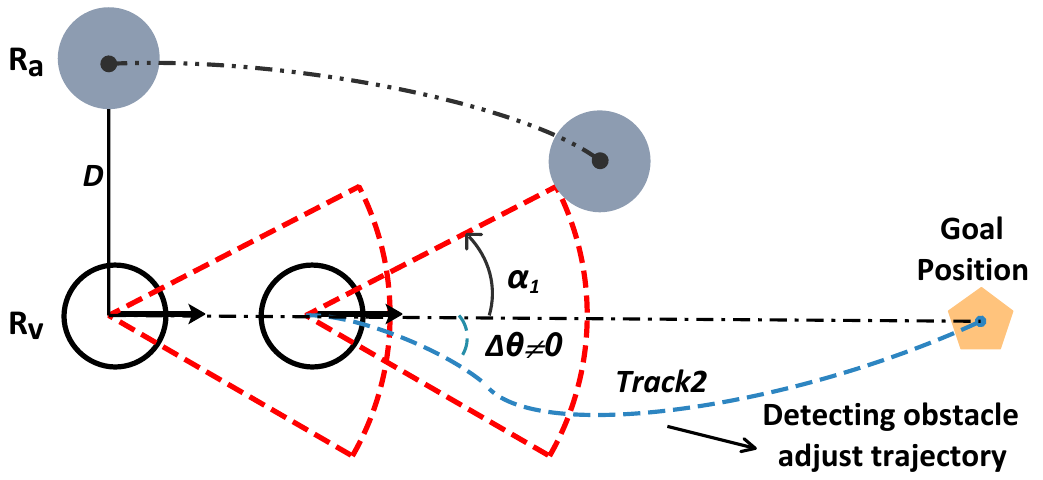} 
  }
  \caption{Illustration of the learning process for $\text{R}_\text{a}$ the . 
  (a) $\text{R}_\text{v}$'s reaction after detecting $\text{R}_\text{a}$ as an obstacle. 
  (b) Learning $\text{R}_\text{v}$'s detection radius $D$. 
  (c) Learning $\text{R}_\text{v}$'s detection ang $\alpha_1$ ($\alpha_2$ is learned likewise at the other side).} 
  \label{Algorithm_1}
\end{figure*}

Note (\ref{eq1}) is a noise-free formulation and the real observation of $\text{R}_\text{a}$ are corrupted with Gaussian noise $\mathcal{N}(0,\sigma^2)$. 
According to the $3\sigma$-principle \cite{pukelsheim1994three}, for Gaussian distribution with zero expectation value, the variable locates in $[-3\sigma,+3\sigma]$ with probability 0.997, and this interval is basically regarded as the actual range. 
Therefore, we take $3\sigma$ as the maximum error bound of the corresponding measurement in this paper. 
And the analysis of the measurement noise will be given later. 

For $\text{R}_\text{a}$, it also needs to discriminate what type $\text{R}_\text{v}$ is first. 
This can be done by checking the smoothness of the reaction trajectory, for the velocity direction of the holonomic robot is easy to change suddenly while that of the non-holonomic robot is strictly constrained. 
Based on (\ref{eq1}), $\text{R}_\text{a}$ can master $\text{R}_\text{v}$'s instantaneous motion information of any time. 
This constitutes the foundation of the following steps. 

\subsection{Intentional Excitation: Trajectory Trial}
The obstacle detection area of $\text{R}_\text{v}$ can be generalized to a circular region mostly. 
If we do not consider obstacles behind the robot when it moves forward, the detection area is modeled as a sector directly. 
Then, at this stage, the primary objective of $\text{R}_\text{a}$ is to infer the radius $D$ and angle range $\bm{\alpha}$ of the sector, which are together denoted as $(D,\bm{\alpha})$.

When $\text{R}_\text{a}$ moves close to $\text{R}_\text{v}$, $\text{R}_\text{a}$ makes a record of $\text{R}_\text{v}$'s relative position, heading and bearing with it. 
After a period $\tilde T$, $\text{R}_\text{a}$ does the measurement again. 
With the two readings, $\text{R}_\text{a}$ is able to calculate $\text{R}_\text{v}$'s position variation $\Delta s$ and heading variation $\Delta\theta$ after $\tilde T$, as shown in Fig.~\ref{infer1}. 
During time slot $[t^s_k,t^s_{k+1}]$, we have
\begin{equation}\label{output1}
\left\{
\begin{aligned}
{\Delta s(k)}&=\sqrt{{{(\Delta x_{k,k+1})}^2} +{{(\Delta y_{k,k+1})}^2}},\\
{\Delta \theta(k)}&=\text{arctan}\frac{{\Delta y_{k,k+1}}}{{\Delta x_{k,k+1}}},
\end{aligned}
\right.
\end{equation}
where $\Delta x_{k,k+1}\!=\!x(t^s_{k}+\tilde T)-x(t^s_k)$ and $\Delta y_{k,k+1}\!=\!y(t^s_{k}+\tilde T)-y(t^s_k)$. 
Let $p_g$ represent the goal position of $\text{R}_\text{v}$. 
In normal situations, $\text{R}_\text{v}$ goes straight forward to $p_g$, i.e., $\Delta \theta\!=\!0$. 
Therefore, $\text{R}_\text{a}$ is detected as an obstacle within $(D,\bm{\alpha})$ by $\text{R}_\text{v}$ if $\Delta \theta\ne0$. 
Meanwhile, after the orientation of $\text{R}_\text{a}$ becomes stable, $\text{R}_\text{a}$ also records the trajectories of $\text{R}_\text{v}$ moving straightforward to $p_g$, as shown in Fig.~\ref{infer2} and ~\ref{infer3}.

It is remarkable that when the mission area is obstacle-free, moving straight to the goal position is the most efficient way. 
There are also cases where the robot is specified to go to the goal in a smooth curve, and we can utilize classic polynomial fitting technique to approximate the trajectory \cite{chen2016tracking}. 
Let $n_p$ be the polynomial order, $\bm{\tilde{x}}=[x^{n_p},x^{n_p-1},\cdots, x^{0}]^\mathsf{T}$ and $\bm{P}\in\mathbb{R}^{n_p+1}$ be the coefficient vector associated with $\bm{\tilde{x}}$. 
A point in the trajectory is modeled as 
\begin{equation}\label{eq:poly-equation}
y=\bm{P}^\mathsf{T} \bm{\tilde{x}} + \eta,
\end{equation}
where $\eta$ is i.i.d. Gaussian noise satisfying $N(0,\sigma^2)$.
Then, the problem is transformed to find the best $\bm{P}$ from groups of linear equations (\ref{eq:poly-equation}), where the solving procedures are the same as that of a straight line ($n_t=1$ in this case). 
Therefore, we mainly focus on the latter case for simplicity. 

Repeatedly, suppose there are $n_z$ groups of trajectories recorded, and the observed points associated with $l$-th trajectory satisfy 
\begin{equation}\label{eq:observations}
\mathcal{T}_r^{(l)}:~y_{lj}=\bm{P}_l \bm{\tilde{x}}_{lj} + \eta_{lj},~j=1,2,\cdots, n_t, 
\end{equation}
where $n_t$ is the number of recorded points in this trajectory. 
Let $d_T(\mathcal{T}_{\vartheta},p)$ be the minimal distance from a position $p$ to $\mathcal{T}_{\vartheta}$, and we obtain the following result. 

\begin{theorem}[Estimate the goal position]\label{th-0001}
Given $n_t$ observed points in $\mathcal{T}_r^{(l)}$, if ${n_t} \to \infty$, the estimated $\mathcal{T}_r^{(l)}$ will pass through $p_g$, satisfying
\begin{equation}
\mathop {\lim }\limits_{{n_t} \to \infty } d_T(\mathcal{T}_r^{(l)}, {p_g}) = 0.
\end{equation}
Given trajectories $\{\mathcal{T}_r^{(l)}, l=1,\cdots, n_z \}$ and let $\hat{p}_g(n_z)$ be the corresponding least square estimator of the goal position $p_g$. If ${n_z \to \infty }$, $p_g$ is accurately estimated, i.e., 
\begin{equation} \label{th-0001-2}
\mathop {\lim }\limits_{n_z \to \infty } {\left\| {{\hat p}_{g}(n_z) - {p_g}} \right\|_2} = 0.
\end{equation}
\end{theorem}
\begin{proof}
The proof is provided in Appendix \ref{pr:th-0001}. 
\end{proof}

\begin{remark}\label{remark:rate}
Theorem \ref{th-0001} illustrates how the accuracy is influenced by the recorded trajectories, where $n_t$ determines how close $p_g$ is to a recorded trajectory while $n_z$ determines how close $p_g$ is to $\hat p_g$. 
Specifically, according to Chebyshev inequality (as shown in the proof of Theorem \ref{th-0001}), the estimation errors of a single trajectory $\mathcal{T}_r^{(l)}$ and $\hat{p}_g$ go to zero at the rate of $\bm{O}(\frac{1}{n_t})$\footnote{For two real-valued functions $f_1$ and $f_2$, $f_1(x)=\bm{O}(f_2(x))$ as $x\to x_0$ means $\mathop {\lim }\nolimits_{x \to x_0 } |f_1(x)/f_2(x)|<\infty$.} and $\bm{O}(\frac{1}{n_z})$, respectively. 
\end{remark}

If there are other obstacles occurring in $\text{R}_\text{v}$'s detection area, an efficient way for $\text{R}_\text{v}$ is adding up the impacts of the obstacles and using the composite impact as the instantaneous recorded input. 
Meanwhile, the attack process of $\text{R}_\text{a}$ will simultaneously take $\text{R}_\text{v}$ and other obstacles into account. 
Considering the possibility that $\text{R}_\text{a}$'s sensor view may be blocked by other obstacles, $\text{R}_\text{a}$ can be equipped with advanced onboard sensors at an appropriate height, e.g., deploying an omnidirectional camera on the top of $\text{R}_\text{a}$. 
Also, $\text{R}_\text{a}$ can actively move to other positions such that $\text{R}_\text{v}$ is observed, as in the intentional learning stage. 


\begin{algorithm}[t]
	\small
    \caption{Excitation-based Obstacle-avoidance Learning}
    \label{algo-1}
	\begin{algorithmic}[1] 
	\REQUIRE{Posture regulating variables $\Delta d$, $\Delta \alpha$, constant $k_{\max}$, $\mathcal{J}=\emptyset$}\;
	\ENSURE{Detection area $(D,\bm{\alpha})$, $p_g$ and $f$}\;
    \STATE \textbf{Initialize}: $\text{R}_\text{a}$ moves to remote posture $p_a$ such that $\text{R}_\text{a}$ is directly ahead of $\text{R}_\text{v}$ and $\varphi_r=0$\;
    \WHILE {$\Delta\theta=0$}
    {
       \STATE $\text{R}_\text{a}$ moves towards $\text{R}_\text{v}$ such that $d(p_a,p_v)=d(p_a,p_v)-\Delta d$, and calculate $\Delta\theta$ by (\ref{output1}) \; 
    }
	\ENDWHILE
	\STATE Record $D$, and the following trajectory as $\mathcal{T}_r^{(1)}$. Then reset $\text{R}_\text{a}$ such that $d(p_a,p_v)=D$ and $\varphi_r=+90\degree$
	\WHILE {$\Delta\theta=0$}{
	    \STATE $\text{R}_\text{a}$ moves towards $\text{R}_\text{v}$ such that $\varphi_r=\varphi_r-\Delta \alpha$, and calculate $\Delta\theta$ by (\ref{output1})\;
	}
	\ENDWHILE
	\STATE Record $\alpha_1=\varphi_r$, and following trajectory $\mathcal{T}_r^{(2)}$. Reset $\text{R}_\text{a}$ such that such that $d(p_a,p_v)=D$ and $\varphi_r=-90\degree$. Then $\text{R}_\text{a}$ does the same process again to obtain a new $\alpha_2$ and $\mathcal{T}_r^{(3)}$\;
	\STATE $\bm{\alpha}=[\alpha_2,\alpha_1]$, and use the trajectories to compute $p_g$ by least square method\;
    \STATE Reset $\text{R}_\text{a}$ such that $\text{R}_\text{a}$ is outside $(p_v, D, \bm{\alpha})$\;
    \FOR {$k \gets 1$\ \textbf{to} $k_{\max}$}
    {
      \STATE $\text{R}_\text{a}$ randomly moves to $p_a\in(p_v, D, \bm{\alpha})$, and compute $Q_{in}\!=\![\theta'$, $v_1$, $v_2$, $a$, $d_{va}$, $\varphi_r]$ at $t_k^s$;
      \STATE Wait for a time slot $\tilde T$, compute $Q_{out}=[\Delta s, \Delta \theta]$ at $t^s_{k+1}$, and $\mathcal{J}=\mathcal{J}\cup\{Q_{in},Q_{out}\}$;
    }
    \ENDFOR
    \STATE Use $\mathcal{J}$ and learning-based method (e.g., SVR) to regress $f$\;
    \STATE \textbf{Return} $(D,\bm{\alpha})$, $p_g$ and $f$. 
 	\end{algorithmic}
\end{algorithm}

\subsection{Model learning: SVR-based method}
When $(D,\bm{\alpha})$ of $\text{R}_\text{v}$ is obtained, $\text{R}_\text{a}$ is controlled to excite $\text{R}_\text{v}$ by blocking in $(D,\bm{\alpha})$. 
In this step, $\text{R}_\text{a}$ also records its relative distance $d_{va}$ and bearing $\varphi_r$ with $\text{R}_\text{v}$, and $\text{R}_\text{v}$'s heading deviation  $\theta'$ with ${p_g}$. 
Once $\text{R}_\text{a}$ is detected by $\text{R}_\text{v}$, $\text{R}_\text{a}$ stores two groups of data during the next period $\tilde T$. 
The data groups are defined as  
\begin{equation}\label{input1}
\left\{
\begin{aligned}
Q_{in}(k)&=[\theta'(k), v_1(k), v_2(k), a(k), d_{va}(k), \varphi_r(k)]^\mathsf{T},\\
Q_{out}(k)&=[\Delta s(k), \Delta\theta(k)]^\mathsf{T}. 
\end{aligned}
\right.
\end{equation}
Accordingly, the obstacle-avoidance knowledge dataset $\mathcal{J}$ is constructed by 
\begin{equation}\label{ss}
\begin{aligned}
\mathcal{J} = \left\{ {\mathop  \cup \limits_{k \in \mathcal{F}} \{Q_{in}(k),Q_{out}(k)\} } \right\},\\
\end{aligned}
\end{equation} 
where $\mathcal{F} = \left\{ {k \in {\mathbb{N}^+ }:d_{va}(k) \le D,\varphi _r(k) \in \bm{\alpha} } \right\}$. 

Based on above analysis, the whole steps of intentional learning is designed as Algorithm \ref{algo-1}, by which $\text{R}_\text{a}$ learns the obstacle-detection area $(p_v, D, \bm{\alpha})$ and goal position $p_g$ [Line 2-10], and regresses obstacle-avoidance mechanism $f$ of $\text{R}_\text{v}$ [Line 11-16], respectively. 
The complexity of Algorithm \ref{algo-1} is simply determined by the number of the required observations, where $\text{R}_\text{a}$ just moves to obtain the required independent observations and does not involve other specific computation procedures. 
Taking $p_g$ as an example, two independent groups of observations ($n_z=2$) on two non-parallel trajectories ($n_t=2$) towards $p_g$ are sufficient to obtain an accurate $\hat{p}_g$ in the ideal noise-free situation. 
As Remark \ref{remark:rate} indicates that the estimation error will decay to zero at a rate of $\bm{O}(\frac{1}{n_t})$ in noisy situations, 
setting $n_z=3$ and $n_t=30$ would be sufficient for our purpose, which is also verified in simulations. 
If the position and motion parameter settings change, one only needs to sample more observations to ensure a reliable $\hat p_g$. 


SVR method has good performance on non-linear regression and strong generalization ability when the amount of data isn't vast. 
It is insensitive to the models of the objects and has a certain tolerance for data noises, due to the $\epsilon$-error tube design \cite{cristianini2000introduction}. 
Thus it is adopted in this paper to learn the obstacle-avoidance mechanism using the collected data, which can be formulated as solving the following problem 
\begin{equation}\label{eq:SVR_problem}
\begin{aligned}
&\mathop{\min }\limits_{\omega,\beta, \tilde{\xi}_{k}^{-}, \tilde{\xi}_{k}^{+}}~ \frac{1}{2}\|w\|^{2}+C\sum_{k=1}^{|\mathcal{J}|}\left(\tilde{\xi}_{k}^{-}+\tilde{\xi}_{k}^{+} \right) \\
&\quad \text{s.t.}~~f(Q_{in}(k))-Q_{out}(k) \leq \epsilon+\tilde{\xi}_{k}^{-}, \\
&\quad\quad\quad Q_{out}(k)-f(Q_{in}(k)) \leq \epsilon+\tilde{\xi}_{k}^{+}, \\
&\quad\quad\quad f(Q_{in}(k))=w^\mathsf{T}\psi(Q_{in}(k)) +\beta,\\
&\quad\quad\quad \tilde{\xi}_{k}^{-} \geq 0, \tilde{\xi}_{k}^{+} \geq 0, ~k=1,2,\cdots,|\mathcal{J}|,
\end{aligned}
\end{equation}
where $C>0$ is the regulazation parameter for the $\epsilon$-sensitive loss function, $\tilde{\xi}_{k}^{-}$ and $\tilde{\xi}_{k}^{+}$ are two slack variables, $\psi(Q_{in}(k))$ represents the mapping feature vector of $Q_{in}(k)$ and can be dealt by the commonly used kernel tricks. 
The obtained $\omega$, $\beta$ and the selected kernels together constitute the final learned obstacle-avoidance mechanism $\hat{f}$. 
More details about the solving process of the SVR can be found in \cite{cristianini2000introduction}.


It is worth noting that SVR method utilizes a $\epsilon$-tube as a regularization technique to improve the model robustness, which acquires certain tolerance for the observation noises. 
Also, since the observation noise of $Q_{out}$ accumulates during the observation period $\tilde{T}$, the learning method will achieve the ideal performance if $\tilde T \to T$. 
In addition, although the learned $\hat{f}$ is a mapping from $Q_{in}$ to $Q_{out}$, the model is not necessarily a surjective or injective mapping. 
In other words, $\hat{f}$ cannot be inversely used to obtain the input given outputs in most cases.

The discrepancy of regression for different obstacle-avoidance algorithms embodies in the internal structure and parameters of the regression model. 
With the obstacle-avoidance mechanism learned by $\text{R}_\text{a}$, 
it fills the gap between the strong assumption (where the attacker has known the information of the target system from the very beginning), and the real implementation (where the attacker needs to acquire the necessary information first).

\section{Attack Strategy Design and Analysis}\label{attack-strategy}
In this section, we propose the shortest-path attack and hands-off attack strategies to achieve the final purpose. 
First, the feasibility of the attack strategies is analyzed. 
Then, we present the detailed design of the strategies, 
Finally, the attack performance bounds and conditions are derived to illustrate the effectiveness of our proposed methods. 

\subsection{Attack Idea and Feasibility}

In this part, we demonstrate the ideas of the proposed attack strategies and analyze their feasibility. 

Since path distance is a commonly-used optimization objective in robot path planning \cite{liu2018survey}, we first propose the shortest-path attack strategy for $\text{R}_\text{a}$, which aims to drive $\text{R}_\text{v}$ into the preset trap with minimal path cost. 
Note that $\text{R}_\text{a}$ is in a consistently active state during the shortest-path attack process, which is not desirable in some cases. 
For example, long-time obstacle occurrence nearby may cause an anomaly alarm of $\text{R}_\text{v}$ and even make $\text{R}_\text{v}$ alter the control strategy. 
To alleviate these effects and make $\text{R}_\text{a}$ stealthier, we further propose hands-off attack strategy. 
``Hands-off'' means the control effort over the plant can be maintained exactly at zero over a time interval and aims to obtain the sparsest control or the minimal active interference with the plant \cite{donkers2014minimum,nagahara2015maximum}. 

The feasibility of our proposed strategies lies in two aspects. 
First, the proposed attack is launched from the physical world by making $\text{R}_\text{a}$ disguise as an obstacle in front of $\text{R}_\text{v}$, and the presence of $\text{R}_\text{a}$ is taken into consideration by $f$, making $\text{R}_\text{v}$ unable to ignore the influence. 
Second, the learning scheme proposed in the last section enables $\text{R}_\text{a}$ to acquire the obstacle-avoidance knowledge of $\text{R}_\text{v}$, 
providing reliable support for a more efficient and intelligent strategy design. 
Specifically, the major issue is to address the coupled dynamics of $\text{R}_\text{a}$ and $\text{R}_\text{v}$, and the nonanalytic property brought by the learned $\hat f$.

Based on the learned knowledge about $f$, the obstacle detection area of $\text{R}_\text{v}$ is formulated as 
\begin{align}
&\!\!\mathcal{A}_d(p_v)\!=\! \left\{ {p \!: \! {{\left\| {p - p_v } \right\|}_2} \!<\! D,  \alpha_1\!\le\!\langle\overrightarrow{p p_v},\mathcal{T}_m(p_v) \rangle\! \le\! \alpha_2} \right\}, 
\end{align}
where $\langle\overrightarrow{p p_v},\mathcal{T}_m(p_v) \rangle$ is the angle between the vector $\overrightarrow{p p_v}$ and the motion orientation $\mathcal{T}_m(p_v)$. 

Regardless of the robot type, the motion updates of $\text{R}_\text{v}$ and $\text{R}_\text{a}$ are represented as  
\begin{align}
{p_v}(k + 1) = h_v\left({p_v}(k),u_v(k)\right),\label{at-0}\\
{p_a}(k + 1) = h_a\left({p_a}(k),u_a(k)\right), \label{at-3}
\end{align}
where $u_v(k)$ is the velocity control input of $\text{R}_\text{v}$ and $u_a(k)$ is attack input of $\text{R}_\text{a}$. 
With learned $\hat f$, the velocity input and position of $\text{R}_\text{v}$ are respectively predicted by 
\begin{align}
&\hat u_v(k) = \hat f({p_a}(k),{p_v}(k),u_v(k - 1)), \label{at-1} \\
&{{\hat p}_v}(k + 1) = h_v\left({p_v}(k),\hat u_v(k)\right). \label{at-2}
\end{align}

\subsection{Shortest-path and Hands-off Attack Algorithms}

First, we introduce the shortest-path attack. 
Since the obstacle-avoidance process under $f$ is a dynamic trade-off between avoiding the obstacle and going to goal, there exists a certain moment, before which the obstacle-avoidance part plays a dominant role and after which the go-to-goal does. 
The key point of the shortest-path attack strategy is to impose a consistent impact on the former one as large as possible, such that $\text{R}_\text{v}$ is generally dominated by $\text{R}_\text{a}$ into $p_t$. 

Denote the initial attack position as $p_{a_0}$, and the attack path is the trajectory of $\text{R}_\text{v}$ that starts from $p_{a_0}$ and ends at $p_{t}$. 
Then, to obtain the minimal attack path length, the original attack design problem $\textbf{P}_\textbf{0}$ is rewritten as the following problem. 
\begin{subequations}\label{eq-pro}
\begin{align}
\textbf{P}_\textbf{1}:
\mathop{\min }\limits_{H,{\bm{u}_{a,0:H}}}&C_s(\bm{u}_{a,0:H})=\sum\limits_{k = 0}^H {{\left\| {{{\hat p}_v}(k + 1) - {p_v}(k)} \right\|_2}},\label{eq-pro-a}\\
\text{s.t.}~~~&{\left\| {u_a(k)} \right\|_2} \le \mu, \label{eq-pro-b}\\
&{\left\| {{p_v}(H) - p_t} \right\|_2} \le \delta,\label{eq-pro-c}\\ 
&{\eta \le\left\| {{p_a}(k) - {p_v}(k)} \right\|_2},\label{eq-pro-d}\\
&{p_a}(k) \in \mathcal{A}_d({p_v}(k)),\label{eq-pro-e}\\
&(\ref{at-0}),~(\ref{at-3}),~(\ref{at-1}),~\text{and}~(\ref{at-2}), \nonumber
\end{align}
\end{subequations}
where $\mu$, $\delta$, and $\eta$ are all positive constants, and the constraints hold for all $k \in \left\{ {0, \cdots ,H} \right\}$. 
In (\ref{eq-pro}), the first constraint (\ref{eq-pro-b}) demonstrates the boundness of the attack inputs, and the second constraint (\ref{eq-pro-c}) guarantees that $\text{R}_\text{v}$ is close enough to $p_t$ when the attack stops. 
The $\eta$ in (\ref{eq-pro-d}) is designed to keep a safe distance between $\text{R}_\text{a}$ and $\text{R}_\text{v}$ to avoid possible collision at next move, and (\ref{eq-pro-e}) makes sure that $\text{R}_\text{a}$ is always in the detection area during the attack.  
The last four constraints capture the state dynamics of $\text{R}_\text{a}$ and $\text{R}_\text{v}$. 

\begin{figure}[t]
\centering
\includegraphics[width=0.42\textwidth]{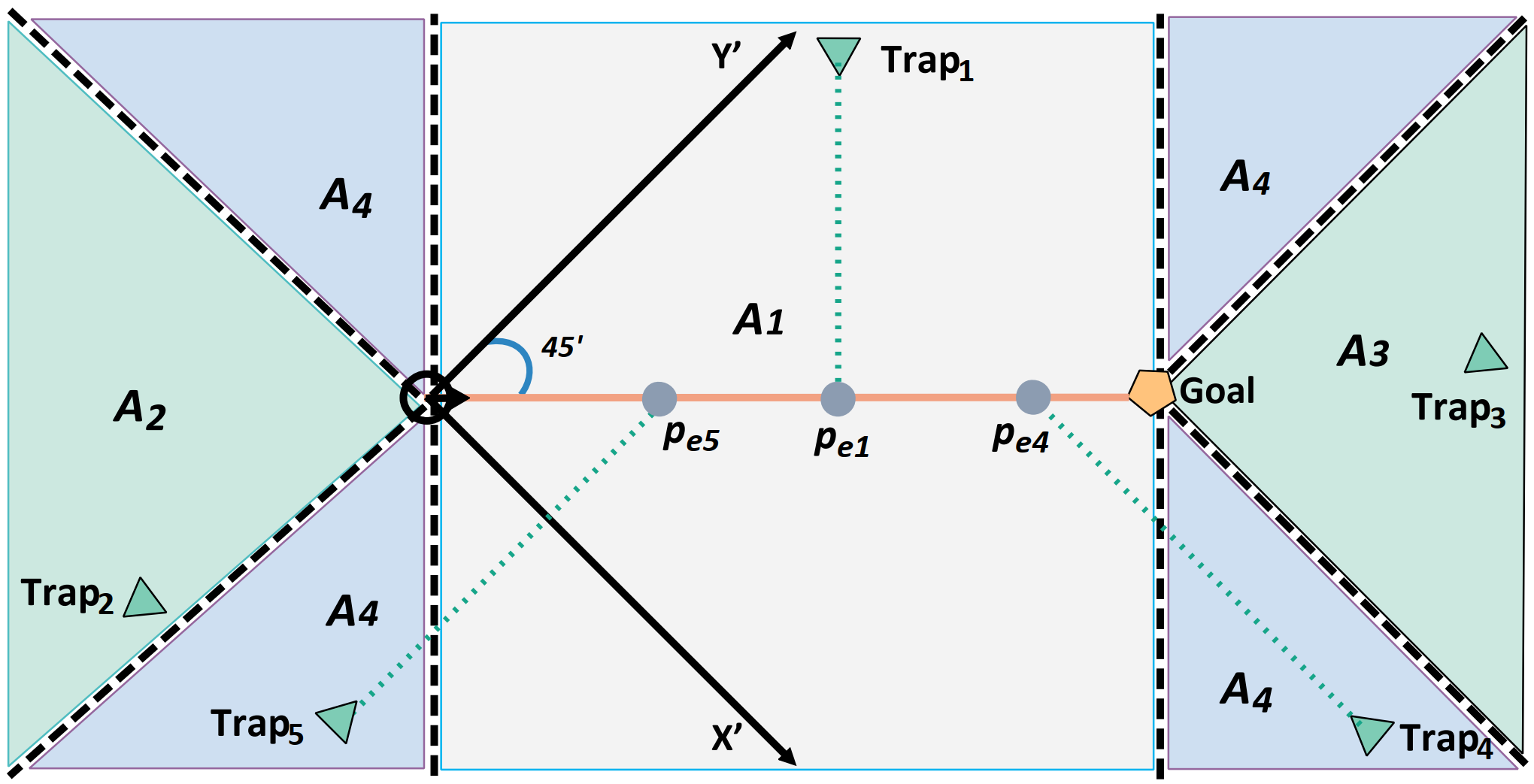}
\caption{Illustration of the choice for entry point. 
To save the page space, we rotate the X'-Y' coordinate frame $45\degree$ clockwise. 
The X'-Y' plane is divided into four parts using dash lines and different colors: the gray area between two vertical dash lines is $\mathcal{A}_1$, the green area between the two left clinodiagonal dashed lines is $\mathcal{A}_2$, the green area between the two right clinodiagonal dashed lines is $\mathcal{A}_3$, and the left blue region is $\mathcal{A}_4$. 
The entry point of $Trap_1$, $Trap_4$ and $Trap_5$ is $p_{e1}$, $p_{e4}$ and $p_{e5}$, respectively.  }
\label{fig-th2-1}
\vspace*{-5pt}
\end{figure}
Ideally, the shortest attack path $\mathcal{T}_a$ is directly obtained by connecting ${p_{{a_0}}}$ and $p_t$. 
However, $\mathcal{T}_a$ is almost impossible to be the real optimal trajectory, because the mechanical structure and control dynamics constraints of $\text{R}_\text{v}$ will not allow the sudden turn point and the trajectory rigidity in $\mathcal{T}_a$. 
Even so, we are able to take $\mathcal{T}_a$ as evaluation criteria: the closer the obtained attack trajectory is with 
$\mathcal{T}_a$, the better the attack strategy is. 
Then comes an interesting problem: how to choose the best $p_{a_0}$ to begin attacking, 
which we call entry point. 


Denote by $\mathcal{T}_g$ the trajectory from the initial position $p_v(0)$ to $p_g$ without being attacked. 
Then, the definition of the entry point is given as follows.

\begin{definition}\label{defi1}
\emph{(Entry point)} 
Given a configuration of $p_v(0)$, $p_g$, and $p_t$, 
rebuild a coordinate frame X'-Y' where the Y' positive axis is at a angle of $45\degree$ to $\mathcal{T}_g$. 
Divide the X’-Y’ plane into four parts using dash line (as shown in Fig.~\ref{fig-th2-1}). 
Then, the entry point $p_e$ is defined as follows
\begin{enumerate}[i)]
\item If $p_t \in \mathcal{A}_1$, $p_e$ is the vertical projection of $p_t$ onto $\mathcal{T}_g$, i.e,  
\begin{equation}\label{eq-pe}
{p_e} = \arg \mathop {\min }\limits_{p'} {\rm{ }}{\left\| {p' - p_t} \right\|_2^2},
\end{equation}
where $p' \in \{p: I_T(\mathcal{T}_g,p)=0\}$. 
\item If $p_t \in \mathcal{A}_2$, there is no exact position of $p_e$. 
Under given constraints, the earlier to attack, the better. 
\item If $p_t \in \mathcal{A}_3$, there is no exact position of $p_e$. 
Under given constraints, the later to attack, the better. 
\item If $p_t \in \mathcal{A}_4$, $p_e$ is the projection of $p_t$ along X' or Y' direction onto $\mathcal{T}_g$. 
\end{enumerate}
\end{definition}

Given a preset small threshold $r_d\!>\!0$ (e.g., it can be set as the smallest position deviation between two observation moments), if $\| p_v(k)-{p_e} \|_2\le r_d$ holds for the first time during $\text{R}_\text{v}$'s movement towards $p_g$, then set $a_s=k$. 
Next, $\text{R}_\text{a}$ begins attacking by moving to its initial attack position $p_a(a_s+1)$, which satisfies 
\begin{equation}\label{cond-2}
\left\{
\begin{aligned}
&{p_a}({a_s} + 1)\!\in\! \mathcal{A}_d({p_v}({a_s})),\\
&I_T(\mathcal{T}_g,p_a(a_s+1))\!\cdot\! I_T(\mathcal{T}_g,p_t)<0.
\end{aligned}
\right.
\end{equation}
Note the second condition in (\ref{cond-2}) requires the initial attack position and $p_t$ are in opposite positions of $\mathcal{T}_g$.

\begin{algorithm}[t]
	\small
    \caption{Shortest-path Attack Strategy}
    \label{algo-3}
    \begin{algorithmic}[1]
    \REQUIRE{$\hat f$, the iteration limit $k_{\max}$, and tolerant bound $\delta$.}
    \ENSURE{Terminal horizon $H$, and attack input vector ${\bm{u}_{a,0:H}}$};

    \STATE Initialize $p_v(0)$, $p_g$, $p_t$, and set the attack signal $a_s=0$;
    \STATE Compute entry point $p_e$ by (\ref{eq-pe});
    \STATE Randomly select a $p_a(0)$ such that $ \| p_a(0)-p_e \|\le r_d$;
    \STATE $\text{R}_\text{v}$ starts to run towards $p_g$, $\mathcal{A}_a^f=\emptyset$;
    \FOR {$k \gets 1$\ \textbf{to} $k_{\max}$}
    {
    	\IF {$a_s=0$ and $ \| p_v-p_e \|\le r_d$}
		{
			\STATE $\text{R}_\text{a}$ moves into the pre-attack position, $a_s=k$;
		}
	  	\ELSE
	  	{
	  		\STATE $\text{R}_\text{a}$ stays still;
	  	}
	  	\ENDIF
	  	\IF {$a_s\ge1$}
	  	{
	  		\STATE Uniformly sample a subset $\mathcal{A}_a^{\text{rand}}\subseteq\mathcal{A}_d(p_v(k))$;
	  		\FOR{$p_a(k+1) \in \mathcal{A}_a^{\text{rand}}$}
	  		{
	  			\STATE Compute $\hat p_v(k+1)$ by (\ref{at-2}); 
	  			\STATE $\hat d_{vt}(k \!+\! 1) \!=\! {\left\| {{\hat p_v}(k \!+\! 1) \!-\! p_t} \right\|_2}$, $d_{vt}(k) \!=\! {\left\| {{p_v}(k) \!-\! p_t} \right\|_2}$;
		    	\IF {$\hat d_{vt}(k+1)\!\le\! d_{vt}({k})$ and $p_a(k+1)\! \in \! \mathcal{A}_d(p_v(k))$}
				{
					\STATE Update $\mathcal{A}_a^f\!=\!\mathcal{A}_a^f  \cup \{ p_a(k+1)\}$;
				}
				\ENDIF
	  		}
	  		\ENDFOR
	  		\STATE $k'\!=\!k\!-\!a_s$, $u_a(k') \!=\! \mathop {\arg }\limits_{u_a} \min \{ {\left\| {{\hat p_v}({k} \!+\! 1) \!-\! p_t} \right\|_2}: {p_a(k\!+\!1)} \!\in\! \mathcal{A}_a^f$, (\ref{eq-pro-b}) and  (\ref{eq-pro-d}) hold \};
	  	}
	  	\ENDIF

    	\IF{${\left\| {{p_v}({k}) - p_t} \right\|_2}<\delta$}
		{
			\STATE break;
		}
		\ENDIF
    }
    \ENDFOR
    \STATE $H=k'$, and construct attack input vector ${\bm{u}_{a,0:H}}$;
    \end{algorithmic}
\end{algorithm}

Since the intentional learning and attack process are separated (i.e., there is no direct feedback between the two parts), 
and due to the black-box characteristic of the learned model, obtaining a global analytical solution for the problem is intractable.
Therefore, we propose a sampling-based approach and find sub-optimal solutions quickly. The complete process is summarized in Algorithm \ref{algo-3}.

Algorithm~\ref{algo-3} is composed of three parts: 
i) before $\text{R}_\text{v}$ comes near to $p_e$ (determined by the distance threshold $r_d>0$), $\text{R}_\text{a}$ needs to wait for the best attack time [Line 6-10]; 
ii) after $\text{R}_\text{a}$ begins its attack, every iteration is based on sampling to explore the motion space of both $\text{R}_\text{a}$ and $\text{R}_\text{v}$ and select a best attack input from a feasible attack set $\mathcal{A}_a^{\text{rand}}$ [Line 11-21]; 
iii) when $\text{R}_\text{v}$ is close enough from $p_t$ (determined by the distance threshold $\delta>0$), $\text{R}_\text{a}$ stops and we call the attack is successful [Line 22-24]. 
Note that the complexity of Algorithm~\ref{algo-3} is determined by multiple factors, including the moving offset of $\text{R}_\text{v}$, the distance between ${p_e}$ and ${p_t}$, and the granularity of $\mathcal{A}_a^{\text{rand}}$, making it hard to be explicitly characterized. 
Specially, if $\text{R}_\text{a}$ can make $\text{R}_\text{v}$ directly move towards the trap with constant offset $L$ and by referring to the complexity analysis of gradient-decent algorithm, the complexity of Algorithm~\ref{algo-3} can be characterized by $\bm{O}(\frac{\left\|p_{e}-p_{t}\right\|_{2}^{2}}{\delta L})$ (see Chapter 8.3 in \cite{chong2004introduction} for more proof details). 
This complexity can be regarded as the ideal case where the obstacle-avoidance mechanism is perfectly learned. 
Despite the intractable complexity analysis for general cases, we can characterize the more concerned performance bounds, which will be provided in Section \ref{sub:attack_performance}.


Based on the formulation in the shortest-path attack, we further propose the hands-off attack, which leverages inertia nature that the obstacle-avoidance of $\text{R}_\text{v}$ is not a one-time effort (i.e., it takes certain time to circumvent the obstacle and adjust to its goal position). 
Therefore, $\text{R}_\text{a}$ does not need to move consistently. 
In this scenario, $\text{R}_\text{a}$ is at a hands-off state if it stays still at one moment during the attack. 
To obtain the minimal attack activity period, $\textbf{P}_\textbf{0}$ is rewritten as the following hands-off attack design problem. 
\begin{subequations}\label{eq-pro2}
\begin{align}
\textbf{P}_\textbf{2}:~~~~
\mathop {\min }\limits_{H,{\bm{u}_{a,0:H}}}~&C_h(\bm{u}_{a,0:H})={\left\| {{\bm{u}_{a,0:H}}} \right\|_0} \label{eq-pro2-a}\\
\text{s.t.}~~~~&{\left\| {u_a(t)} \right\|_2} \le \mu, \label{eq-pro2-b}\\
&{\left\| {{p_v}(H) - p_t} \right\|_2} \le \delta, \label{eq-pro2-c}\\
&{\eta_1\le \left\| {{p_a}(t) - {p_v}(t)} \right\|_2} \le \eta_2, \label{eq-pro2-d}\\
&(\ref{at-0}),~(\ref{at-3}),~(\ref{at-1}),~\text{and}~(\ref{at-2}), \nonumber
\end{align}
\end{subequations}
where the constraints hold $\forall t \in \left\{ {0, \cdots ,H} \right\}$. 
In (\ref{eq-pro2}), different from (\ref{eq-pro-d}) and (\ref{eq-pro-e}), (\ref{eq-pro2-d}) can be seen as their relaxation such that $\text{R}_\text{a}$ does not have to be in the obstacle detection area all the time. 
Other constraints are the same as that in $\textbf{P}_\textbf{1}$. 

\begin{algorithm}[t]
	\small
    \caption{Hands-off Attack Strategy}
    \label{algo-4}
    \begin{algorithmic}[1]
    \REQUIRE{$k_{\max}$, $\hat f$, $\delta$ (as in Algorithm \ref{algo-3}), and preset constant $\gamma$;}
    \ENSURE{$H$ and ${\bm{u}_{a,0:H}}$};
    \STATE do [Line 1-4] of Algorithm \ref{algo-3};
    \FOR {$k \gets 1$\ \textbf{to} $k_{\max}$}
    {
	  	\STATE $d_{vt}(k) = {\left\| {{p_v}(k) - p_t} \right\|_2}$, $k'=k-a_s$;
	  	\STATE do [Line 6-10] of  Algorithm \ref{algo-3};
	  	\IF {$a_s\!\ge\!1$ and $d_{vt}(k)>\gamma  d_{te} $}
	  	{
	  		\IF{$d_{vt}(k)\le d_{vt}(k-1)$ and (\ref{eq-pro2-b}), (\ref{eq-pro2-d}) hold}
	  		{
	  			\STATE $\text{R}_\text{a}$ stays inactive, i.e, $u_a({k'})=0$;
	  		}
	  		\ELSE
	  		{
	  			\STATE do [Line 12-20] of Algorithm \ref{algo-3} and obtain $u_a({k-a_s})$;
	  		}
	  		\ENDIF
	  	}
	  	\ELSIF{$a_s\!\ge\!1$ and $d_{vt}(k)\le\gamma d_{te}$}
	  	{
	  		\IF{$d_T(\mathcal{T}_m(p_v(k)),p_t)<\delta$ and (\ref{eq-pro2-b}), (\ref{eq-pro2-d}) hold}
	  		{
	  			\STATE $\text{R}_\text{a}$ stays inactive, i.e, $u_a({k'})=0$;
	  		}
	  		\ELSE
	  		{
	  			\STATE do [Line 12-20] of Algorithm \ref{algo-3} and obtain $u_a(k')$;
	  		}
	  		\ENDIF
	  	}
	  	\ENDIF
	  	\STATE do [Line 22-24] of  Algorithm \ref{algo-3};
    }
    \ENDFOR
    \STATE $H=k'$, and construct attack input vector ${\bm{u}_{a,0:H}}$;
    \end{algorithmic}
\end{algorithm}

The algorithm design of solving $\textbf{P}_\textbf{2}$ is summarized in Algorithm \ref{algo-4}. 
Note that $\text{R}_\text{a}$ cannot stay inactive all the time, thus the major difference from the shortest-path attack here is to maximize the silent period between two consecutive attacks. 
To tackle this issue, we introduce a criteria based on $d_T(\mathcal{T}_m(p_v),p_t)$ and $d_{vt}(k)=d({p_v}(k),p_t)$. 
Then, Algorithm \ref{algo-4} can be interpreted as four parts: i) $\text{R}_\text{a}$ waits for the best attack time first [Line 4], which is similar to Algorithm \ref{algo-3}; 
ii) When $d_{vt}(k)$ is large after the attack begins, $\text{R}_\text{a}$ uses the change of $d_{vt}(k)$ to determine whether to stay still or move [Line 6-10]. 
If it needs to make an attack, the attack inputs are sampled from the motion space of both $\text{R}_\text{a}$ and $\text{R}_\text{v}$ and to explore the best one from the feasible attack set; 
iii) When $d_{vt}(k)$ becomes small, $\text{R}_\text{a}$ alters to a new attack decision criteria, by which $\text{R}_\text{a}$ can stay still if $d_T(\mathcal{T}_m(p_v),p_t)$ does not grow larger and the constraints are satisfied [Line 12-16]; 
iv) Finally, the same stopping criteria as that in Algorithm \ref{algo-3} is adopted [Line 18]. 
Since Algorithm \ref{algo-4} also enjoys the descent-wise nature, it shares the same complexity scale as Algorithm \ref{algo-3}. 


\subsection{Attack Conditions and Algorithm Convergence}
Next, we give the conditions to successfully realize the attack by the proposed algorithms, and analyze the convergence. 

\begin{figure}[t]
\centering
\includegraphics[width=0.38\textwidth]{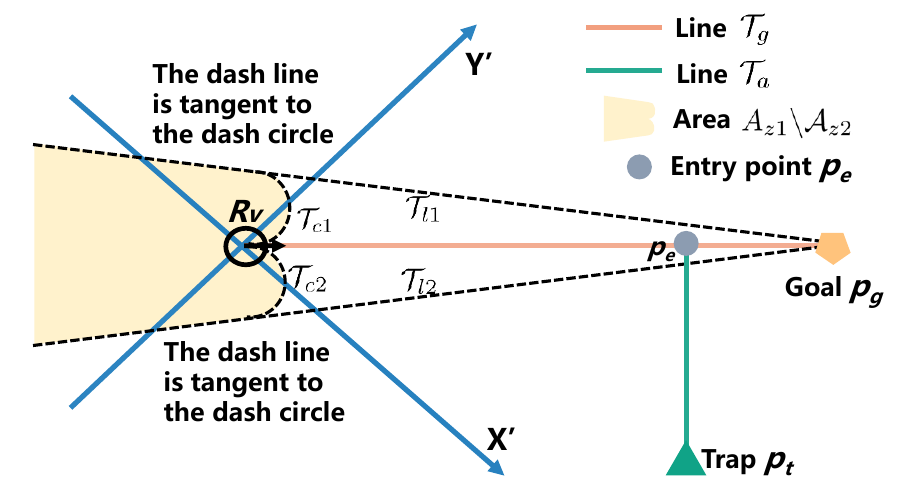}
\caption{Illustration of the condition of successful attack.}
\vspace*{-10pt}
\label{fig-proof2}
\end{figure}

To intuitively describe the trap setting, we mainly use geometric expressions combined with graph illustration. 
Given $p_{v0}$ and $p_g$, 
divide the 2-D plane into the X'-Y' coordinate frame defined in Definition \ref{defi1}. 
First, draw the two circumcircles with radius $r_{\min}$, $\mathcal{T}_{c1}$ and $\mathcal{T}_{c2}$, tangent to $\mathcal{T}_{g}$ at $p_{v0}$, respectively. 
Then, draw the two tangent lines $\mathcal{T}_{l1}$ for $\mathcal{T}_{c1}$, and $\mathcal{T}_{l2}$ for $\mathcal{T}_{c2}$, going through $p_g$. 
In the X'-Y' coordinate frame, with deterministic $p_{v0}$, $p_g$ and $r_{\min}$, the expressions of $\mathcal{T}_{c1}$, $\mathcal{T}_{c2}$, $\mathcal{T}_{l1}$ and $\mathcal{T}_{c1}$ are explicitly determined, and are represented as 
\begin{align}
\mathcal{T}_{c1}&=\{p:\|p-p_{c1}\|_2^2-r_{\min}^2=0\},\\
\mathcal{T}_{c2}&=\{p:\|p-p_{c2}\|_2^2-r_{\min}^2=0\},\\
\mathcal{T}_{l1}&=\{p=(x,y)^T:y-a_1 x-b_1=0\},\\
\mathcal{T}_{l2}&=\{p=(x,y)^T:y-a_2x-b_2=0\},
\end{align}
where $p_{c1}$ and $p_{c2}$ are the centers of $\mathcal{T}_{c1}$ and $\mathcal{T}_{c1}$, and $\{a_1,b_1\}$ and $\{a_2, b_2\}$ are the slope-intercept terms of $\mathcal{T}_{l1}$ and $\mathcal{T}_{l1}$, satisfying $a_1,a_2\!>\!0,b_1\!>\!0$ and $b_2\!<\!0$. 
Accordingly, we have 
\begin{align}
&I_T(\mathcal{T}_{ci},p)=\text{sign}(\|p-p_{ci}\|_2^2-r_{\min}^2), \\
&I_T(\mathcal{T}_{li},p)=\text{sign}(y-a_i x-b_i),
\end{align}
where $i=1,2$ and $\text{sign}$ is the indication function. 
Next, we introduce an auxiliary area variable, given by 
\begin{align}
\!\!\mathcal{A}_{z1}\!=\!\{ p: &I_T(\mathcal{T}_{l1},p)\!\le\!0, I_T(\mathcal{T}_{l2},p)\!\ge\!0, x\!\le\! x^g, y\!\le\! y^g \},
\end{align} 
\begin{align}
\mathcal{A}_{z2}=\{ p: &I_T(\mathcal{T}_{c1},p)\ge0,I_T(\mathcal{T}_{c2},p)\ge0, I_T(\mathcal{T}_{l1},p)\le0, \nonumber  \\
&I_T(\mathcal{T}_{l2},p)\ge0, \|p-p_g\|_2\le\|p_{v0}-p_g\|_2    \}. 
\end{align} 
Apparently, as shown in Fig.~\ref{fig-proof2}, $\mathcal{A}_{z2}\subseteq\mathcal{A}_{z1}$. 
Based on these notations, we first obtain the following theorem to demonstrate the convergence of Algorithm \ref{algo-3}. 

\begin{theorem}[Convergence condition of Algorithm \ref{algo-3}]\label{th:convergence}
If the trap position $p_t \!\notin\! {A}_{z1}\backslash\mathcal{A}_{z2}$, then $\text{R}_\text{a}$ can always drive $\text{R}_\text{v}$ into $p_t$ by Algorithm \ref{algo-3}, i.e., 
\begin{equation}
\mathop {\lim }\limits_{k \to \infty } \| p_v(k) - p_t \|_2 =0. 
\end{equation}
\end{theorem}

\begin{proof}
The proof is provided in Appendix \ref{pr:convergence}. 
\end{proof}

Theorem \ref{th:convergence} presents the criterion to set $p_t$. 
In fact, as Fig.~\ref{fig-proof2} shows, it also demonstrates how the initial configuration of $\text{R}_\text{a}$, $\text{R}_\text{v}$, the goal and the trap will affect the final attack result. 
As long as the initial configurations meet the above conditions, the success of Algorithm~\ref{algo-3} by utilizing the obstacle-avoidance characteristic of $\text{R}_\text{v}$ is guaranteed, which will be also verified with various cases in Section \ref{sec-simulation} and \ref{sec:experiment}.  
There are two points that need to be noted. 
First, although the given condition is only sufficient to achieve the driving-to-trap attack, it covers quite a large area for most trap settings. 
To further relax the condition, the key is to explore more advanced strategy design based on more agile moving ability of $\text{R}_\text{a}$. 
Second, as indicated in the Appendix \ref{pr:convergence}, the obstacle-avoidance characteristic corresponds to a descent property and Algorithm~\ref{algo-3} can be regarded as a descent search algorithm to solve 
\begin{equation}
\mathop{\min }\limits_{ p_v }~~F(p_v)=\left\| p_t - p_v \right\|_2^2,
\end{equation}
which is convex about $p_v$ and has only one minima. 
Note that Algorithm~\ref{algo-3} shares similar form with the conventional gradient-descent method and thus effectively solves the problem. 
The main difference in Algorithm~\ref{algo-3} is that the search direction is not a fixed form at each iteration (e.g., not necessarily the negative gradient of $F(p_v)$). 
By evaluating the most appropriate move from the sampled feasible position set, the algorithm effectively avoids the oscillation issues. 

As for the hands-off attack, Algorithm \ref{algo-4} also works as a descent search like Algorithm \ref{algo-3}, except that the attack robot is not always active. 
However, we note that during an inactive period $[k_1,k_2]$ for the attacker, the positions of the victim satisfy $ \|p_v(k_2)-p_t\|_2 < \| p_v(k_1)-p_t \|_2$, illustrating that the victim still takes a descent path to the trap. 
Hence, the victim's movement during this period is equivalent to a special single step, and the convergence of Algorithm \ref{algo-4} is still guaranteed as Algorithm \ref{algo-3}.

\begin{figure*}[t]
\begin{center}
\subfigure[
]{\label{fig-illustr}
\includegraphics[width=0.23\textwidth]{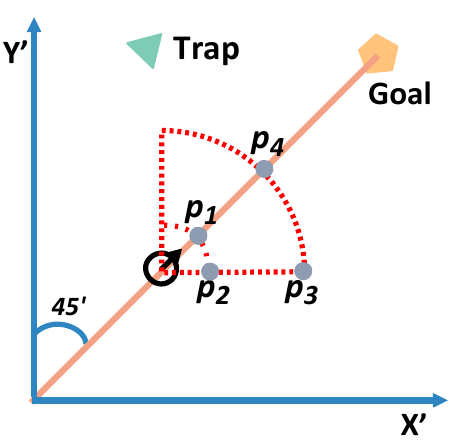}} 
\hspace{10ex}
\subfigure[
]{\label{pattern1}
\includegraphics[width=0.23\textwidth]{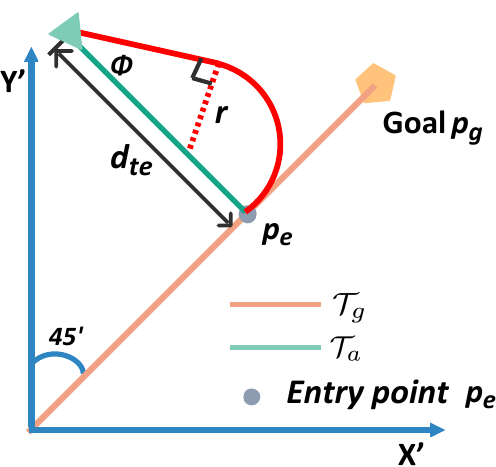}}
\hspace{10ex}
\subfigure[
]{\label{pattern2}
\includegraphics[width=0.23\textwidth]{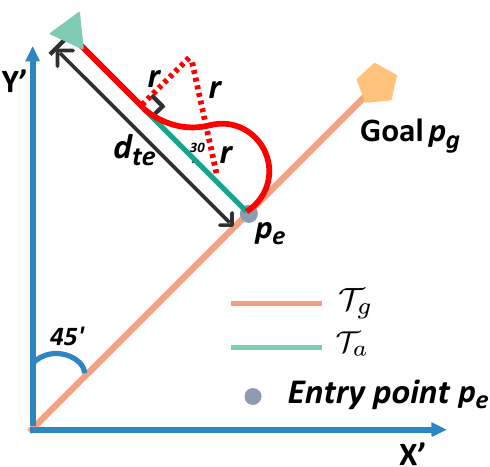}}
\caption{Shortest-path attack against non-holonomic robot. 
(a) Illustrations of four special attack positions $p_i,~i=1,2,3,4.$ 
(b) Attack pattern1: a circular arc plus a line segment. 
(c) Attack pattern2: two circular arcs plus a line segment. 
}
\label{th-illustr}
\end{center}
\vspace*{-5pt}
\end{figure*}

\subsection{Attack Performance Analysis}\label{sub:attack_performance}

In this part, we analyze the attack performance of Algorithms~\ref{algo-3} and \ref{algo-4} from the perspective of the path length and activity period, respectively. 
First, denote by $C_s^*$ the optimal cost of $\textbf{P}_\textbf{1}$, and we have the following result. 

\begin{theorem}[Existence of $\varepsilon$-approximated solutions]\label{th3}
The optimal cost can be approximated with arbitrary precision, i.e., $\forall \varepsilon>0$, there exists a $\bm{u}_{a,0:H}$ such that $|C_s(\bm{u}_{a,0:H})-C_s^*|\le\varepsilon$. 
\end{theorem}
\begin{proof}
The proof is provided in Appendix \ref{pr:th3}. 
\end{proof}


Theorem \ref{th3} demonstrates the existence of $\varepsilon$-approximated solutions for the optimal cost of $\textbf{P}_\textbf{1}$. 
To further investigate how good the solution by Algorithm \ref{algo-3} is, we introduce a two-tuple $(\omega,r)$ to illustrate the visual effect that how a non-holonomic robot avoids an obstacle
\footnote{For a non-holonomic robot, the output $(v,\omega)$ under $f$ also determines the curvature radius $r$ as $v=\omega r$, and $(\omega,r)$ is more intuitive to describe the instantaneous obstacle-avoidance behavior.}.
As shown in Fig.~\ref{fig-illustr}, supposing the trap is at the left side of $\text{R}_\text{v}$, we denote four extreme positions $\left\{ {{p_1},{p_2},{p_3},{p_4}} \right\}$ where $\text{R}_\text{a}$ could make an effective attack, along with corresponding output denoted as $\left\{ {({\omega _i},r_i),i = 1, \cdots ,4} \right\}$. 
In terms of the reaction radius ${r_i}$, $p_1$ is the most threatening position for $\text{R}_\text{a}$ while $p_3$ the least threatening, as ${r_1} <{r_2}<{r_4}<{r_3}$. 
Let ${r_{\max }} = \max \left\{ {{r_i}} \right\}$, ${r_{\min }} = \min \left\{ {{r_i}} \right\}$, $d_{te}={\left\| {p_t-p_e} \right\|_2}$, 
and denote ${l_{path}}$ as the trajectory length of $\text{R}_\text{v}$ under attack. 
Then, we present the following theorem that gives the performance bounds of Algorithm \ref{algo-3}.
\begin{theorem}[Performance bounds of Algorithm \ref{algo-3}] \label{th004}
Let $\bar C_s$ be the cost obtained by Algorithm \ref{algo-3}. 
When $p_e \in \mathcal{A}_1$ and $d_{te} > 2 \cdot r_{\min}$, we have
\begin{small}
\begin{equation}\label{eq_th1}
(\pi/2+\xi  \!-\! \cos \xi ){r_{\min }} + d_{te}(\cos \xi  - 1) \!\le\! \bar C_s- {C_s^*} \!\le\! (\frac{7}{6}\pi  - 1 - \sqrt 3 ){r_{\max }}, 
\end{equation}
\end{small}
\!\!\!\!~where $\xi = \arcsin ( {\frac{{{r_{\min }}}}{{d_{te} - {r_{\min }}}}} )$.
\end{theorem}
\begin{proof}
The proof is provided in Appendix \ref{pr:th004}. 
\end{proof}

With the feasibility of the attack guaranteed, Theorem \ref{th004} indicates in most situations\footnote{There is no need to consider $p_t\!\in\! \mathcal{A}_1$ such that $d_T(\mathcal{T}_g,p_t)\!\le\! (\sqrt{3}\!+\!1)r_{\min}$ in Algorithm \ref{algo-3}, as the attack can be easily achieved from $p_e$.} 
how close the solution obtained by Algorithm \ref{algo-3} is to ideal trajectory, and how the worst solution could be, which is extremely hard to actually meet.


Concerning the hands-off attack, the existence of the optimal solution of $\textbf{P}_\textbf{2}$ is similar with $\textbf{P}_\textbf{1}$ and guaranteed. 
The proof resembles that of Theorem \ref{th3} and is omitted here.  
Then, the attack performance of Algorithm \ref{algo-4} is presented in the following theorem. 
\begin{theorem}[Performance bounds of Algorithm \ref{algo-4}]\label{th-last}
Let $H_1$ and $H_2$ denote the attack horizon obtained by Algorithm \ref{algo-3} and \ref{algo-4}, respectively, then we have 
\begin{equation}
H_2\le2H_1,
\end{equation}
and the active attack period by Algorithm \ref{algo-4} satisfies 
\begin{equation}
C_h(\bm{u}_{a,0:H_2})/H_2\le 0.5.
\end{equation}
\end{theorem}
\begin{proof}
The proof is provided in Appendix \ref{pr:th-last}. 
\end{proof}
Theorem \ref{th-last} gives the attack performance bounds of Algorithm \ref{algo-4}, in terms of the attack path length (compared with Algorithm \ref{algo-3}) and the attack activity ratio. 
In Section \ref{sec-simulation}, we will illustrate the worst performance is hard to actually meet. 
Note that the core insight of Algorithm 3 lies in how to leverage the inertia property of $\text{R}_\text{v}$'s obstacle-avoidance behavior to make $\text{R}_\text{a}$ more inactive, not in how to design the best attack move for $\text{R}_\text{a}$ when it needs be active. 
Therefore, it can also be compared with other attack algorithms by replacing the [Line 9, 15] of Algorithm 3 with the attack move computed by other algorithms, and the similar conclusions to Theorem 5 still hold.

\subsection{Attack against Holonomic Robots}
For holonomic robots, the attack performance (results of Theorems \ref{th004}-\ref{th-last}) for $\text{R}_\text{a}$ is the same under Algorithm \ref{algo-3} and \ref{algo-4}. 
The major difference is that the motions of holonomic robots in two orthogonal directions are independent and involve no coupling constraints, and there is no need to compute their orientations, making learning the obstacle-avoidance mechanism easier for $\text{R}_\text{a}$. 
Therefore, attacks against holonomic robots with omnidirectional moving ability are not subject to the limitations of Theorem \ref{th3}, i.e., the preset trap is arbitrary. 
However, due to the same reason, a stronger moving ability for $\text{R}_\text{a}$ is required to move to its optimal attack position.

\section{Simulations}\label{sec-simulation}

In this section, we model both non-holonomic and holonomic robots as simulation objects. 
Specifically, we adopt the popular dynamic window approach (DWA) \cite{fox1997dynamic,seder2007dynamic} and artificial potential method (APM) \cite{khatib1986real} as the obstacle-avoidance mechanism of the two kinds of robots, respectively. 
Our evaluation focuses on two aspects: effectiveness and efficiency. 
For better illustrating and depicting the attack process as detailed as possible in a square figure, we set the initial positions of $\text{R}_\text{a}$, $\text{R}_\text{v}$, the trap, and the goal roughly on the four corners (other settings that meet the attack conditions exhibit similar results). 
First, the critical learning steps of the proposed attack framework are shown. 
Then, we illustrate the effectiveness of the proposed control designs by presenting different attack scenarios and evaluating the attack costs. 
Finally, detailed comparisons with other works are provided. 

\begin{figure}[t]
\begin{center}
\subfigure[]{\label{example_1}
\includegraphics[width=0.22\textwidth]{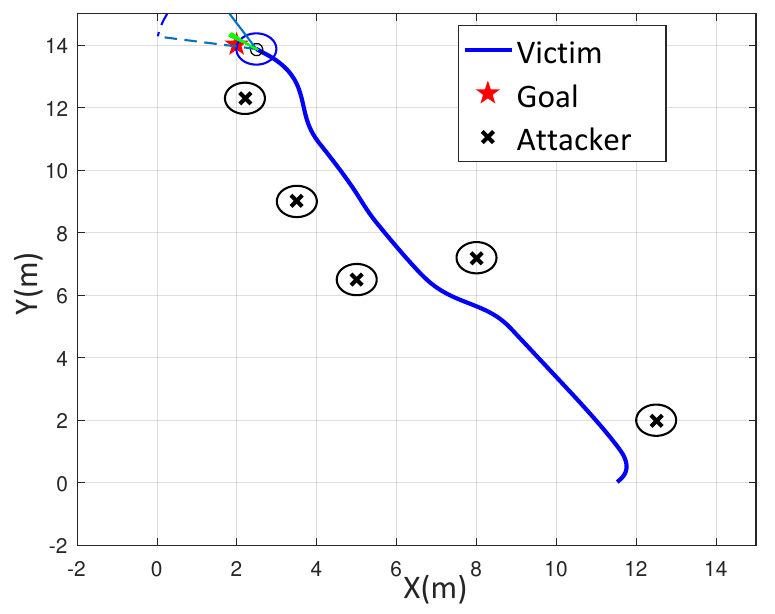}} 
\subfigure[]{\label{simple_attack_1}
\hspace{1.3ex}
\includegraphics[width=0.22\textwidth]{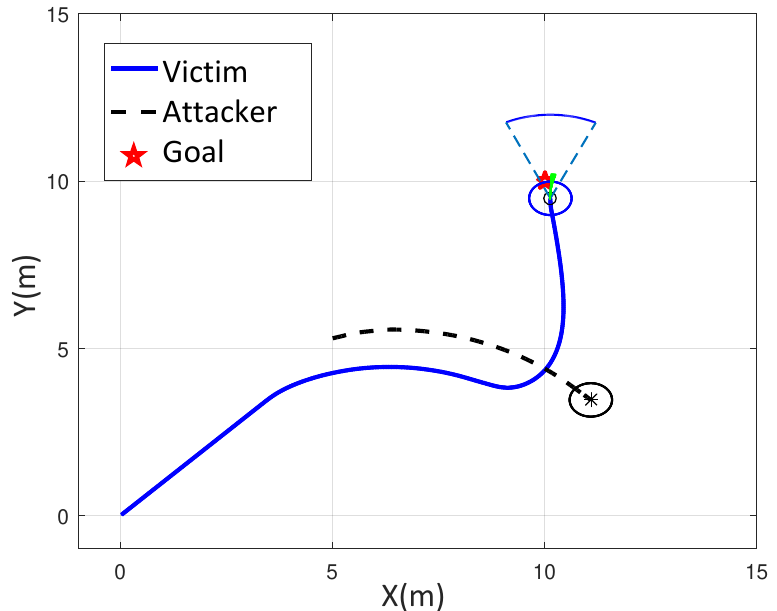}}
\caption{(a) Intentional learning: $\text{R}_\text{a}$ disguises as an obstacle to influence $\text{R}_\text{v}$'s running, and then collects the reaction data of $\text{R}_\text{v}$ at `$\times$' positions consecutively. 
$\text{R}_\text{v}$ moves from $(11.5,0)$ to $(2,14)$. 
(b) Simple attack: once $\text{R}_\text{a}$ appears in $(D, \bm{\alpha})$, it makes continuous impacts on $\text{R}_\text{v}$ in one direction (here is to the right).
}
\label{Intentional Learning}
\end{center}
\end{figure}

\begin{figure}[t]
\begin{center}
\subfigure[]{\label{h-attack0_1}
\includegraphics[width=0.22\textwidth]{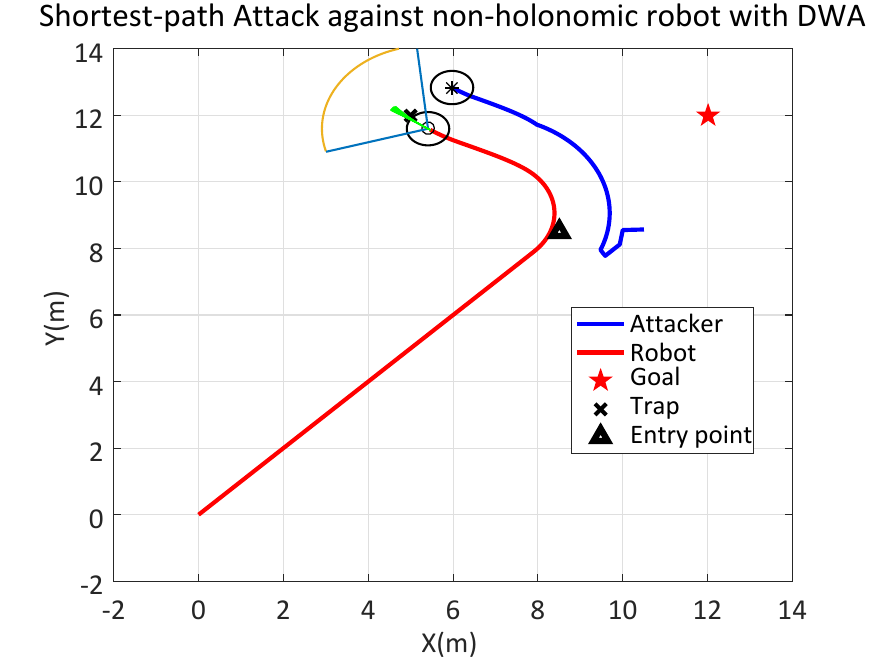}}
\subfigure[]{\label{h-attack0_2}
\includegraphics[width=0.22\textwidth]{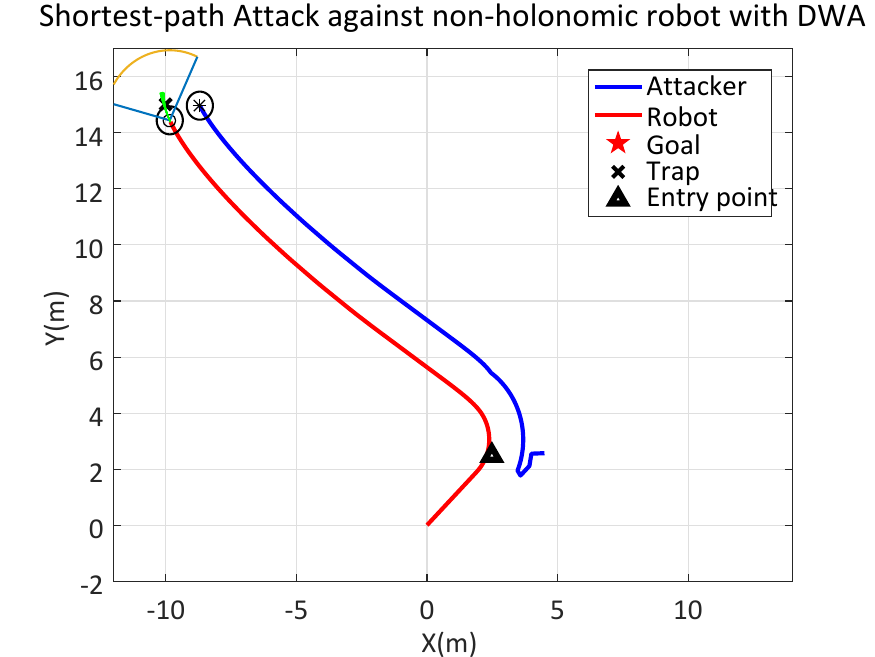}}
\caption{Shortest-path attack against non-holonomic robot. The obstacle-avoidance algorithm of $\text{R}_\text{v}$ is DWA. 
(a) The preset trap locates in $(5,12)$. Attack iteration time is 89 steps, and the path length after being attacked is 5.52m.
(b) The preset trap locates in $(-10,15)$. Attack iteration time is 296 steps, and the path length after being attacked is 18.35m.
}
\label{h-attack0}
\end{center}
\end{figure}

\begin{figure}[t]
\begin{center}
\subfigure[]{\label{h-attack1-1}
\includegraphics[width=0.228\textwidth]{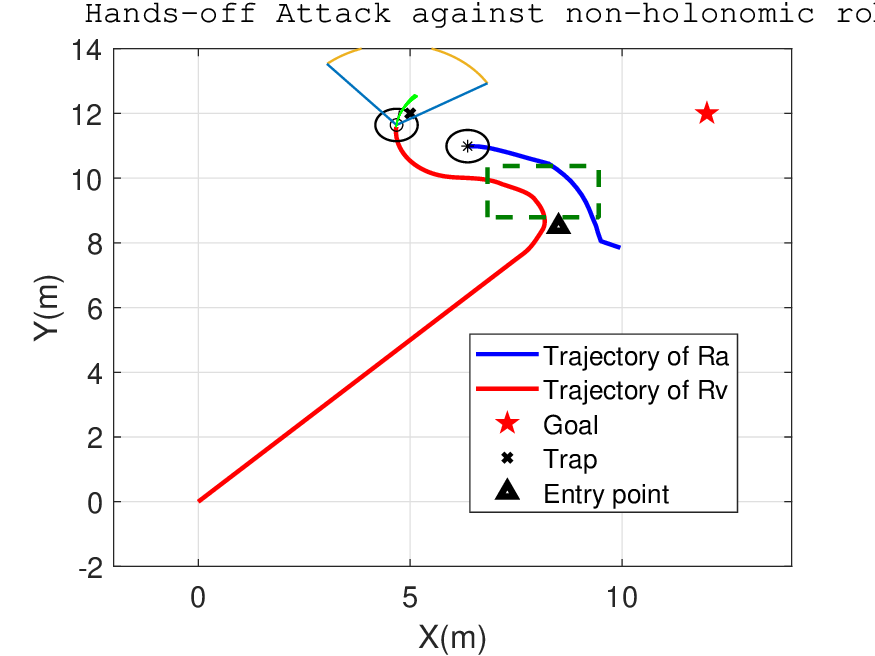}} 
\subfigure[]{\label{h-attack1-2}
\includegraphics[width=0.228\textwidth]{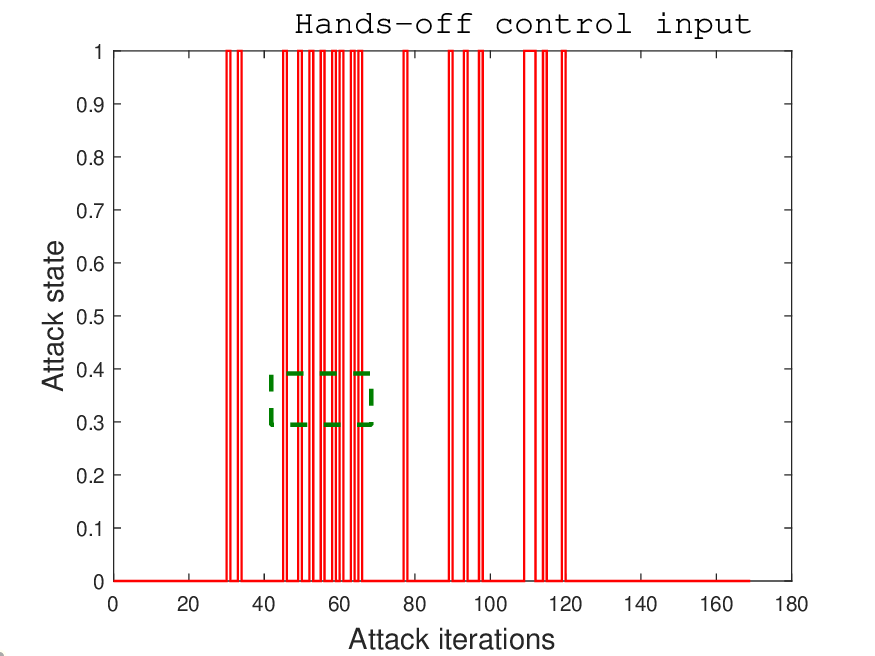}}
\subfigure[]{\label{h-attack2-1}
\includegraphics[width=0.228\textwidth]{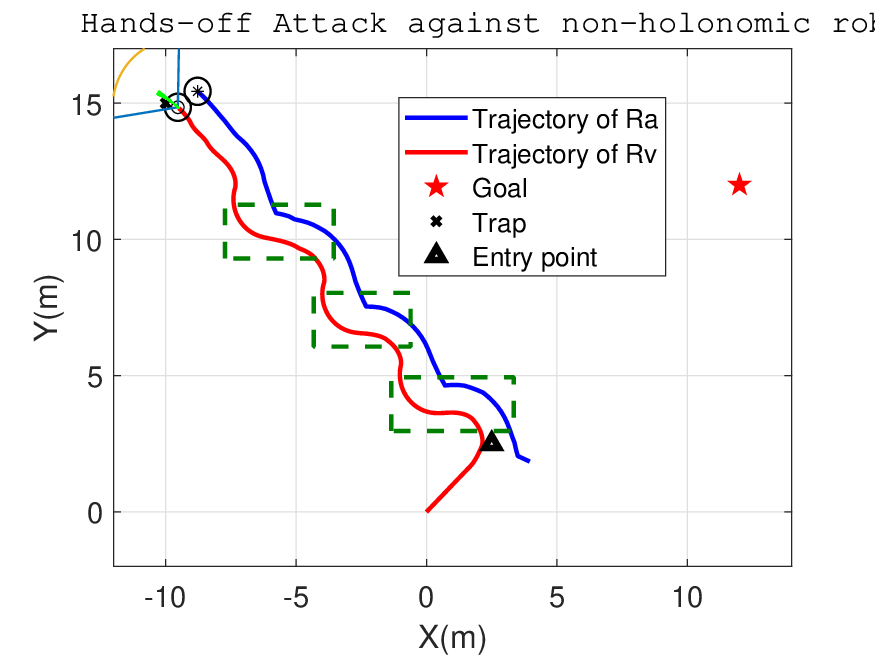}} 
\subfigure[]{\label{h-attack2-2}
\includegraphics[width=0.228\textwidth]{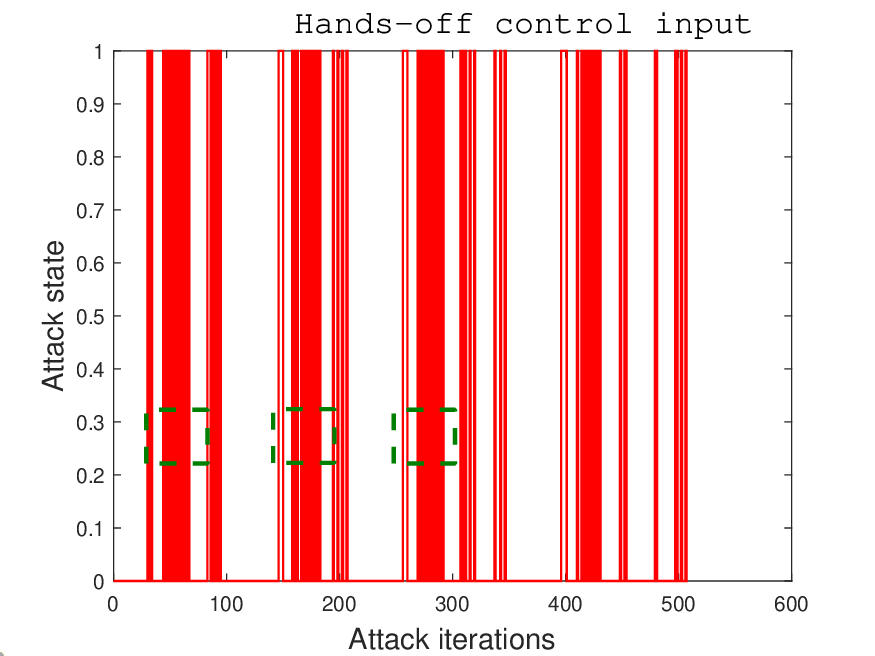}}
\caption{Illustration of the hands-off attack against non-holonomic robot. 
The obstacle-avoidance algorithm of $\text{R}_\text{v}$ is DWA. 
In hands-off control input, the state 1 means $\text{R}_\text{a}$ is active while state 0 means $\text{R}_\text{a}$ stays still. 
The marked parts with green rectangles mean the attack move is intensive. 
(a) and (b): The preset trap locates in $(5,12)$. Attack horizon is 169 steps, $\text{R}_\text{a}$ is active only for 20 steps, and the path length after being attacked is 6.56m.
(c) and (d): The preset trap locates in $(-10,15)$. Attack horizon is 509 steps, $\text{R}_\text{a}$ is active only for 78 steps, and the path length after being attacked is 20.43m. 
}
\vspace*{-5pt}
\label{h-attack1}
\end{center}
\end{figure}

\subsection{Learning Phase}
Fig.~\ref{example_1} shows that $\text{R}_\text{a}$ collects a series of $Q_{in}$ and $Q_{out}$ by sequential ``intentional learning''. 
It is feasible to regress the obstacle-avoidance algorithm using the data, when no prior information of that is available.
Fig.~\ref{simple_attack_1} shows a simple and rough attack: once $\text{R}_\text{a}$ appears in $(D, \bm{\alpha})$, it predicts $\text{R}_\text{v}$'s next move, and runs in the predicted direction with faster speed and repeats this process. 
As expected, $\text{R}_\text{v}$ should keep avoiding $\text{R}_\text{a}$ all the time. 
However, this attack cannot proceed consistently. 
We conclude the leading cause lies in two parts: 
i) The control inputs are not well-designed by considering where the next (sub-)optimal attack position is; 
ii) Each prediction will produce certain inaccuracy, which is cumulated to influence the attack effect in this case. 
Next, we present the results of the proposed attack algorithms. 
By comparison, the effectiveness of the algorithms is exhibited, remedying the deficiency of the former one.

\begin{table*}[t]
  \centering
  \caption{\label{tab:attack_comparison} Comparisons with representative works of herding problems} 
  \begin{tabular}{ccccccc}
  \toprule
 \multirow{2}{*}{ \textbf{Works} }   & \multirow{2}{*}{ {\begin{tabular}[c]{@{}c@{}} \textbf{Dynamic model } \\ \textbf{ of target robot}${^1}$ \end{tabular}}} &  \multirow{2}{*}{ {\begin{tabular}[c]{@{}c@{}} \textbf{Self-goal} \\ \textbf{consideration}${^2}$ \end{tabular}}}  & \multirow{2}{*}{ {\begin{tabular}[c]{@{}c@{}}\textbf{Model is not } \\ \textbf{priorly known}${^3}$ \end{tabular}}}  & \multirow{2}{*}{  {\begin{tabular}[c]{@{}c@{}} \textbf{Make target robot} \\ \textbf{go preset position} \end{tabular}}} & \multirow{2}{*}{  {\begin{tabular}[c]{@{}c@{}} \textbf{Required number} \\ \textbf{of active robots}  \end{tabular}}} \\
  \\
  \midrule
   \cite{pierson2017controlling} &  potential function based &    &   & $\checkmark$  & multiple \\ 
  \midrule
  \cite{varava2017herding} &  potential function based  &  &  partially known & $\checkmark$ & multiple \\
  \midrule
   \cite{licitra2019single} &  nonlinear Lipschitz function  &   & $\checkmark$ & $\checkmark$ & single  \\
  \midrule
  \cite{paranjape2018robotic} &  flocking model  &  &   &  not fixed & single \\
  \midrule
 \cite{chipade2021multiagent} &  flocking model &  &  $\checkmark$ &  $\checkmark$ & multiple \\
  \midrule
  Our work &  \makecell[c]{ majority of obstacle-\\avoidance algorithms} & \makecell[c]{$\checkmark$}  & \makecell[c]{$\checkmark$} & \makecell[c]{$\checkmark$} & \makecell[c]{single} \\
  \bottomrule
  \addlinespace[0.5ex]
  \setlength\tabcolsep{0.5ex}
  \vspace{-8pt}
  \end{tabular}
  \begin{tablenotes}[para]\footnotesize
      \item[1] For comparisons in a uniform framework, the herd robot in herding problem and the victim robot in our work are both referred as the target robot, the herder robot in herding problem and the attack robot in our work are both referred as the active robot. 
      \item[2] This factor indicates that whether the dynamic model of target robot contains an intrinsic goal position, i.e., the dynamics of the target robot is not only determined by the active robot, but also by its intrinsic goal position. 
      \item[3] This means whether the active robot knows the dynamic model of the target robot to design control strategies. 
  \end{tablenotes} 
\end{table*}

\subsection{Attack Phase}
\begin{figure}[t]
\begin{center}
\subfigure[]{\label{APF1}
\includegraphics[width=0.227\textwidth]{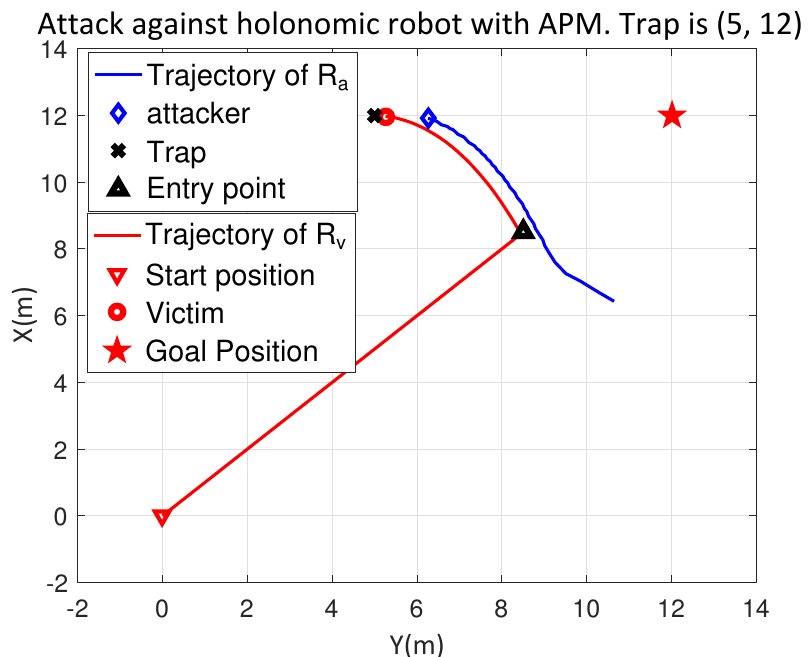}}
\subfigure[]{\label{APF2}
\includegraphics[width=0.227\textwidth]{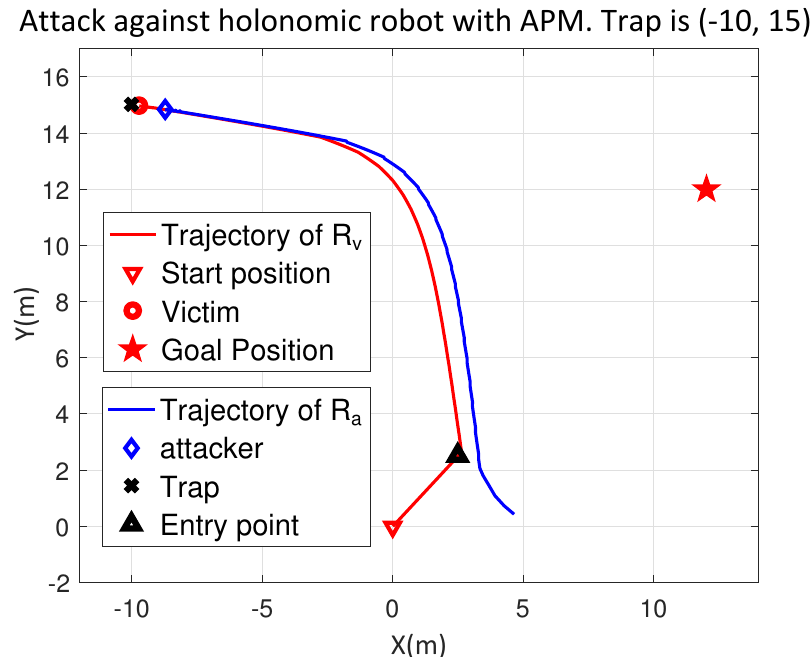}}
\caption{Illustration of the shortest-path attack against holonomic robot. The obstacle-avoidance algorithm of $\text{R}_\text{v}$ is APF. 
(a) The preset trap locates in $(5,12)$. And the path length after being attacked is 5.13m. 
(b) The preset trap locates in $(-10,15)$. And the path length after being attacked is 20.78m.
}
\label{SAttack-APF}
\end{center}
\vspace*{-5pt}
\end{figure}

\textbf{Basic setup}: 
The goal position of $\text{R}_\text{v}$ is set as $(12,12)$. 
For a fair comparison, the same two cases for all attack strategies are designed where the trap is set as $(5,12)$ and $(-10,15)$, respectively. 
The traps locates in area $\mathcal{A}_1$ (defined in Fig.~\ref{fig-th2-1}), satisfying the conditions in Theorem \ref{th3}. 
Fig.~\ref{h-attack0} shows that the shortest-path attack strategy against a non-holonomic robot is applied. 
Fig.~\ref{h-attack1} shows that the hands-off attack strategy against a non-holonomic robot is applied. 
Examples of the holonomic robot are presented in Fig.~\ref{SAttack-APF}. 

\textbf{Results analysis}: 
In terms of path cost, from Fig.~\ref{h-attack0} we can observe that the trajectory of $\text{R}_\text{v}$ is near to $\mathcal{T}_a$ by the shortest-path attack, and the trajectory of $\text{R}_\text{v}$ by hands-off attack twists and turns a little more, as shown in Fig.~\ref{h-attack1}. 
Therefore, the former attack strategy is better in the sense of attack path length. 
For example, it is 5.52m in Fig.~\ref{h-attack0_1} while 6.56m in Fig.~\ref{h-attack1-1}, consistent with the conclusion in Theorem \ref{th-last}. 
However, if we focus on the active attack time, it is found that the small sacrifice of path length largely saves the attack times for $\text{R}_\text{a}$ by hands-off attack. 
As shown in Fig.~\ref{h-attack1-2} and \ref{h-attack2-2}, $\text{R}_\text{a}$ only needs to attack in part times of the whole process to implement the attack successfully.
The marked parts with green rectangle show this part of trajectory twists and turns, meaning the attack moves in this period are intensive. 
The reason lies in that $\text{R}_\text{a}$ needs to make sure that $\text{R}_\text{v}$ moves towards the trap, not adjusting to the goal direction in time [Line 6-21, Algorithm \ref{algo-3}]. 
Taking the case of trap $(-10,15)$ as an example, by hands-off attack, the attack horizon is 509 steps and $\text{R}_\text{a}$ is active only for 78 steps, consistent with Theorem \ref{th-last}.

For a holonomic robot $\text{R}_\text{v}$, 
note that its trajectory is not as smooth as that of a non-holonomic robot, i.e., the moving direction of $\text{R}_\text{v}$ may suddenly change. 
Even though this property makes the movement of $\text{R}_\text{v}$ more unconstrained, 
the response trajectory of $\text{R}_\text{v}$ is not necessarily shorter than that of a non-holonomic robot, as we can see from Fig.~\ref{h-attack0_2} and Fig.~\ref{APF2}. 
The reason is as follows. 
If the obstacle detection range of this obstacle-avoidance model is neglected, we can always obtain a corresponding obstacle position to make $\text{R}_\text{v}$ move to any desired position. 
Nevertheless, in our design which takes real limitations into consideration, the feasible positions of $\text{R}_\text{a}$ to implement an attack are strictly constrained, formulated as (\ref{eq-pro2-d}) or (\ref{eq-pro-d}) and (\ref{eq-pro-e}). 
Therefore, a desired next-step position may not be feasible, and we can only choose a position whose effect is closest to the desired position.

\subsection{Comparisons with Other Works}

The proposed attack shares some similar features with conventional pursuit-evasion problems (e.g., \cite{vidal2002probabilistic,kolling2009pursuit,mejia2019solutions}) and herding problems (e.g., \cite{lee2017autonomous,varava2017herding,licitra2019single}),  
as they all utilize the interaction characteristic of the robots to achieve the specific tasks. 
However, it would be inappropriate to directly compare the method performance with these works, due to the major difference in their basic model formulation and application scenarios. 
For intuitive illustrations, the comparisons with some representative works are provided in Table \ref{tab:attack_comparison}, which can be interpreted from the following three aspects. 
\begin{itemize}
\item \textit{Task scenario.} 
The proposed attack aims to drive the target robot into a preset trap position by utilizing the obstacle-avoidance mechanism. 
While in pursuit-evasion problems, the aim of the pursuers is to capture the evaders, which is usually turned to closely track the evaders or form an encirclement over the evader. 
\item \textit{Model knowledge.} 
Our work focuses on learning the knowledge about the interaction mechanism of mobile robots and then utilizing the learned model to design a driving-to-trap attack, while most existing works rarely discuss the former issue about model knowledge, e.g., many herding problem works assume the interaction model of the robot is priorly known. 
\item \textit{Model types:} 
The proposed attack is not tailored for a specific dynamic model of the victim robot. 
The dynamics of the robots in the pursuit-evasion is generally modeled as a multi-agent LTI systems, and in herding problems it is usually characterized by specific function forms (e.g., potential function), and only depends on the positions of the herding and target robots.
\end{itemize}

\begin{figure}[t]
\centering
\includegraphics[width=0.4\textwidth]{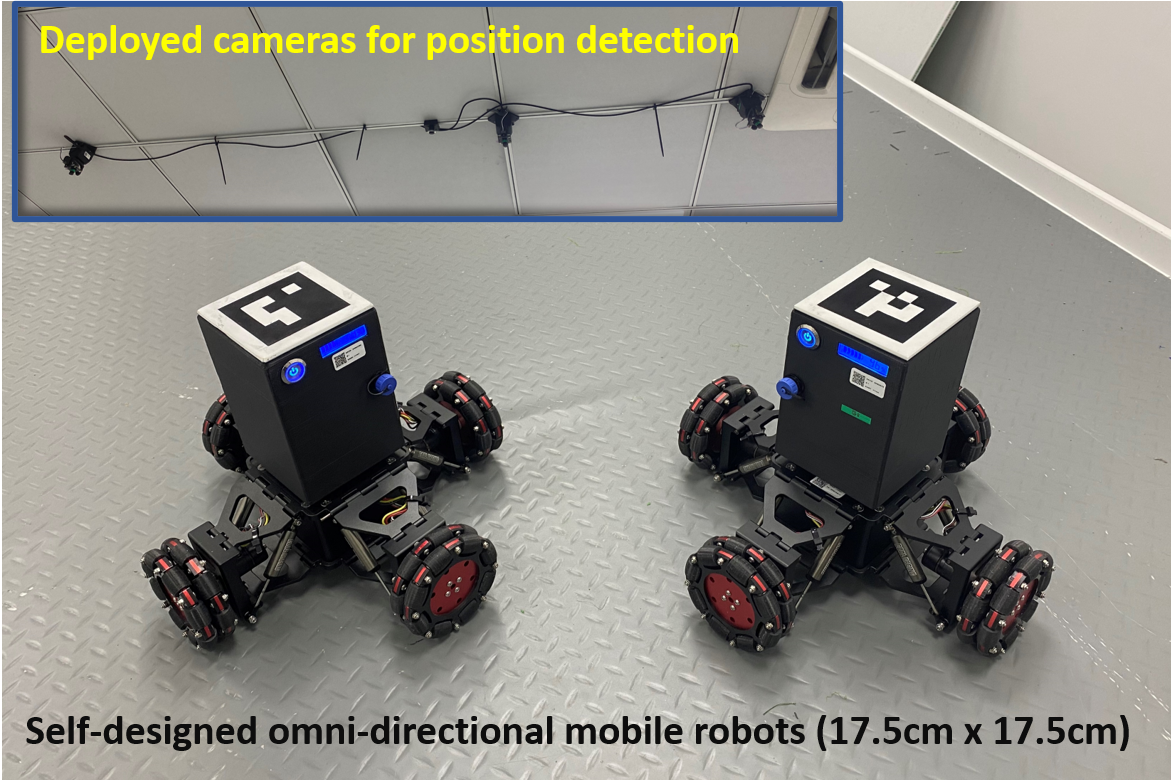}
\caption{The robot platform preview. 
Each robot is equipped with an unique code pattern on its top face for position detection. 
The position detection is performed by the multi-camera system deployed to the laboratory ceiling.}
\vspace*{-10pt}
\label{fig:platform}
\end{figure}


\begin{figure*}[t]
\begin{center}
\subfigure[]{\label{fig:learning2}
\includegraphics[width=0.24\textwidth]{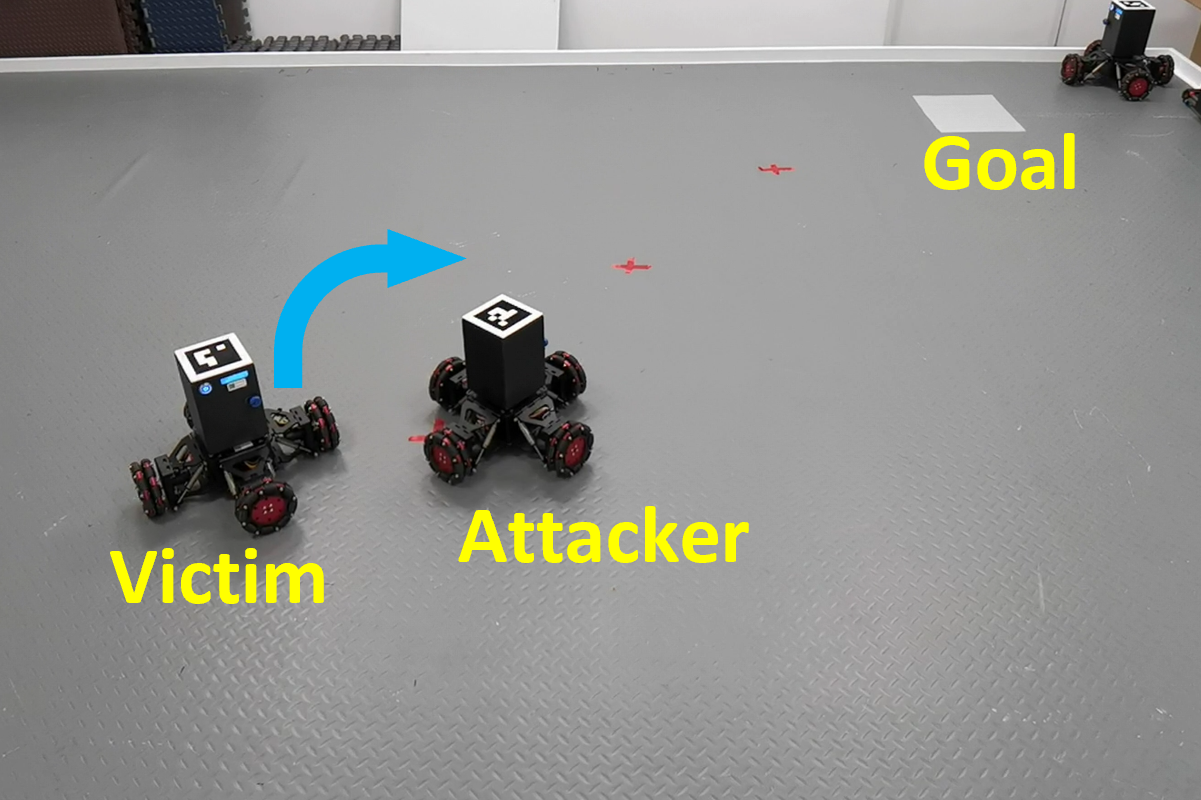}}
\subfigure[]{\label{fig:learning3}
\includegraphics[width=0.24\textwidth]{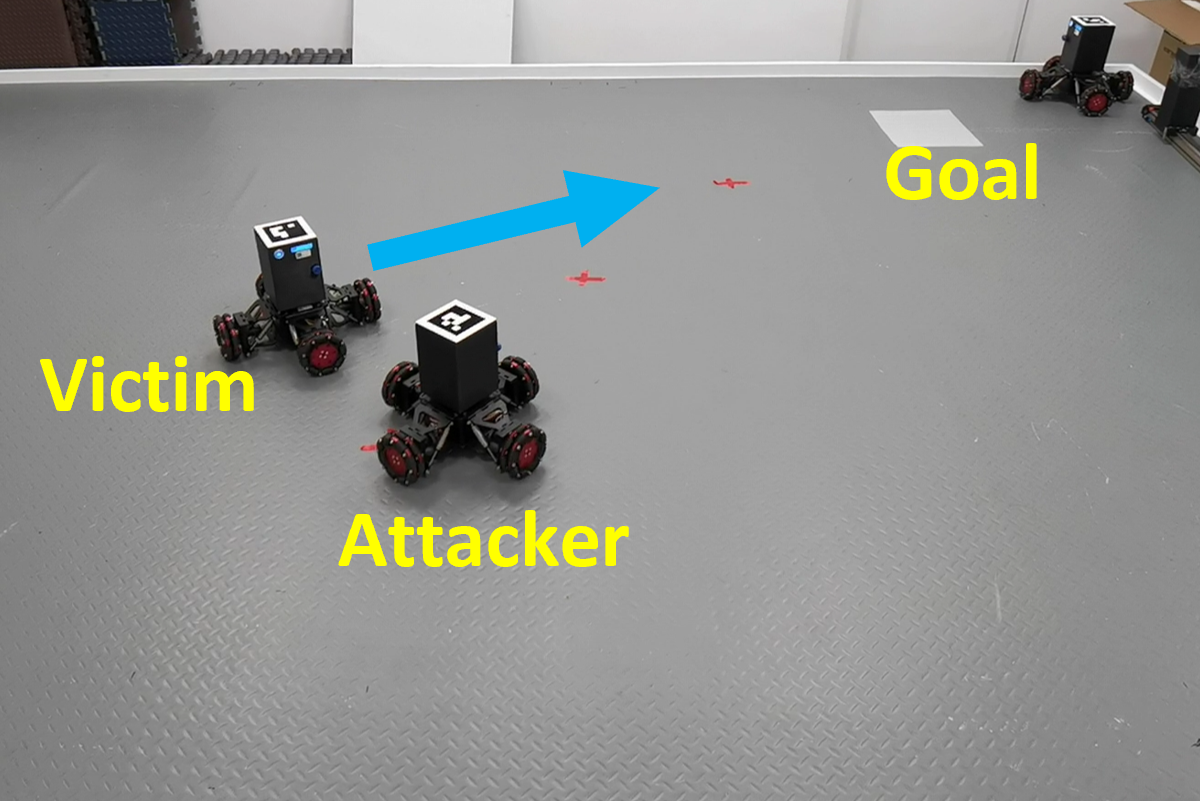}} 
\subfigure[]{\label{fig:learning1}
\includegraphics[width=0.24\textwidth]{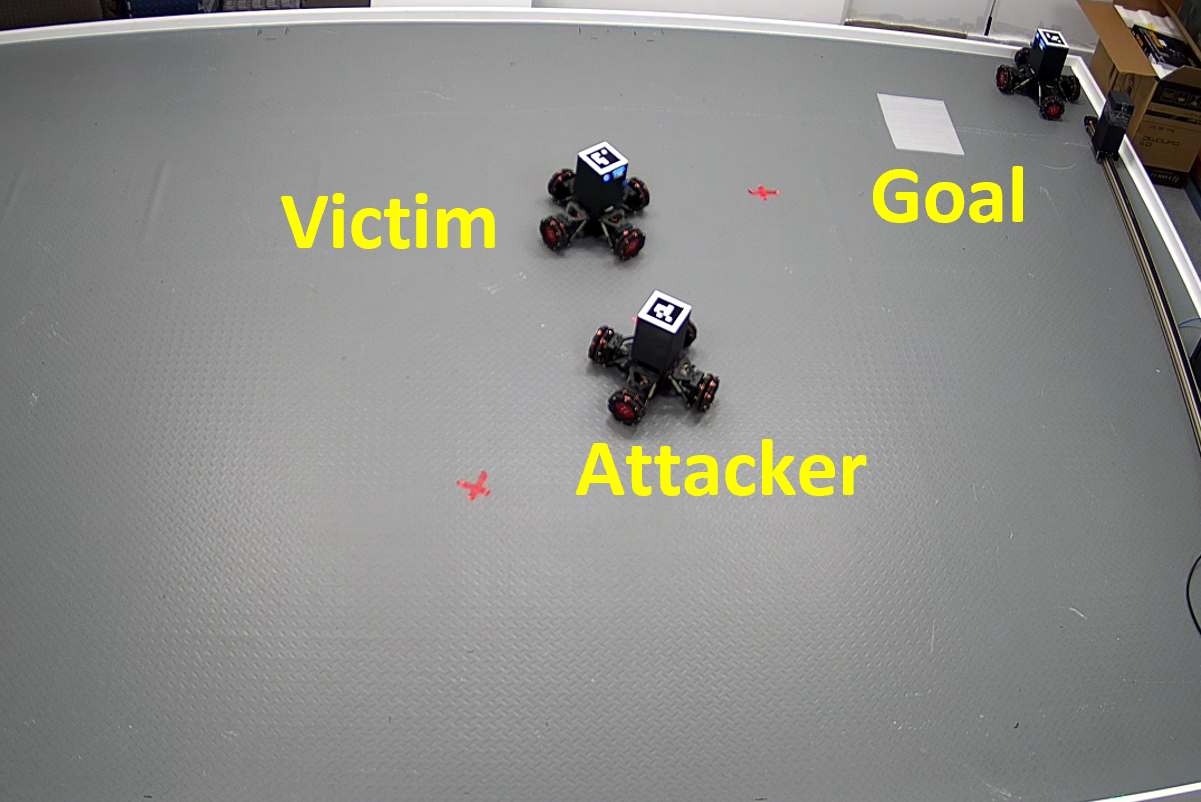}} 
\subfigure[]{\label{fig:learning4}
\includegraphics[width=0.22\textwidth]{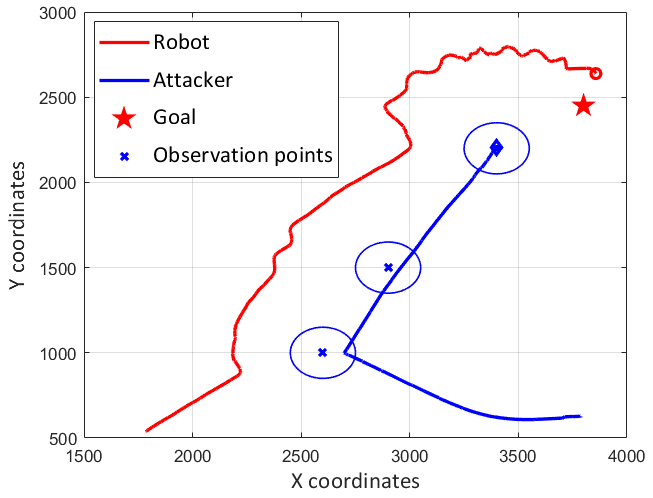}}
\vspace*{-5pt}
\caption{Illustration of the intentional learning process. 
(a) $\text{R}_\text{v}$ detects $\text{R}_\text{a}$ and try to avoid it. 
(b) After $\text{R}_\text{a}$ is off the detection range of $\text{R}_\text{v}$, $\text{R}_\text{v}$ will continue approaching the goal. 
(c) $\text{R}_\text{a}$ moves to the next observation point and makes impacts on $\text{R}_\text{v}$ again. 
(d) The trajectories of the robots in this learning example. 
}
\label{fig:active_learning}
\end{center}
\end{figure*}

\section{Experiments}\label{sec:experiment}

\subsection{Platform Description and Experiment Setup}
To demonstrate the practical performance of the proposed attack, multiple experiments were conducted in our self-designed mobile robot platform \cite{ding2021robopheus}. 
The AprilTag visual system is adopted for the real-time localization of the robots. 
The control procedures based on the localization results are implemented by MATLAB in a VMWare ESXI virtual machine, which is equipped with an Intel(R) Xeon(R) Gold 5220R CPU, 2.20G Hz processor and 16GB RAM.

We validate the result on a $\SI{5}{\meter} \times \SI{3}{\meter}$ rectangular space in practice. 
Fig.~\ref{fig:platform} shows the used robots, which are with $\SI{17.5}{\cm} \times \SI{17.5}{\cm} \times \SI{20}{\cm}$ volumes and designed in omni-directional form. 
The robots are subject to the motion model \eqref{eq:motion-controller-2}, and their real-time positions are detected by the cameras deployed to the lab ceiling. 
The positions of the robot, goal, trap and other obstacles are represented by the coordinates of their geometric central points. 
Note that the unit of all motion states (e.g., position and speed) are described by the picture pixel hereafter, where $\SI{1}{pixel}$ in the calibrated picture is about $\SI{1.08}{\mm}$ in the practical testbed.

For simplicity and intuitive comparison, the same initial position settings of the victim robot $\text{R}_\text{v}$, the attacker $\text{R}_\text{a}$, the goal $p_g$, and the preset trap $p_t$ are used in different types of experiments. 
The initial positions of the victim robot and the attacker are $p_v(0)=[1800,600]$ and $p_a(0)=[3800,600]$, respectively. 
The goal and trap positions are $p_g=[3800,2400]$ and $p_t=[1900,2200]$, respectively. 
According to Definition \ref{defi1}, the entry point for the attacker is $p_e=[2670,1373]$. 
The maximum velocity of the attacker and the victim robot are $\SI{400}{pixel/s}$ and $\SI{200}{pixel/s}$. 
The obstacle-detection radius is set as $D=\SI{600}{pixel}$. 
The robot is assumed to arrive at the goal or trap position, if $\|p_v-p_g\|_2<200$ or $\|p_v-p_t\|_2<200$.

\begin{figure}[t]
\subfigure[]{\label{fig:SA_ob1}
\includegraphics[width=0.23\textwidth]{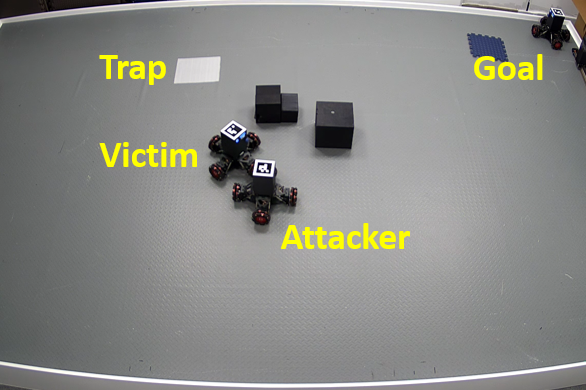}} 
\subfigure[]{\label{fig:SA_ob2}
\includegraphics[width=0.23\textwidth]{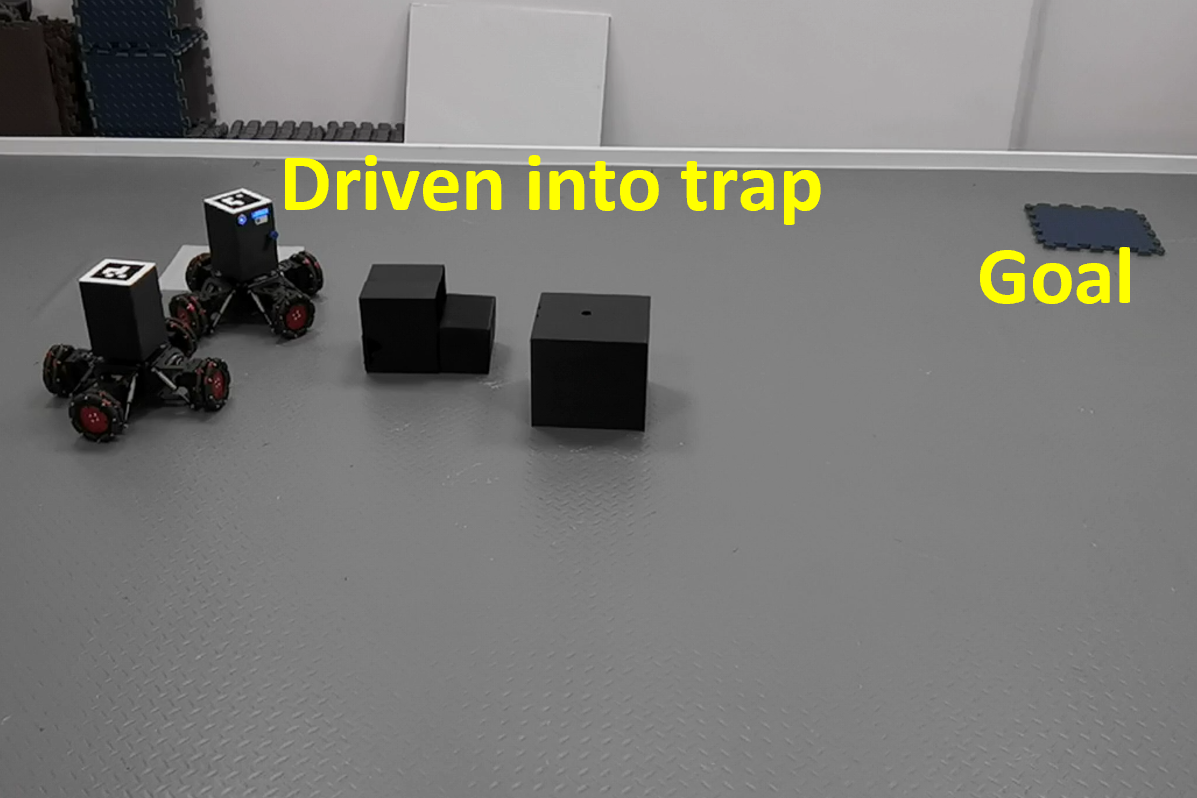}}
\subfigure[]{\label{fig:SA_ob_state}
\includegraphics[width=0.235\textwidth]{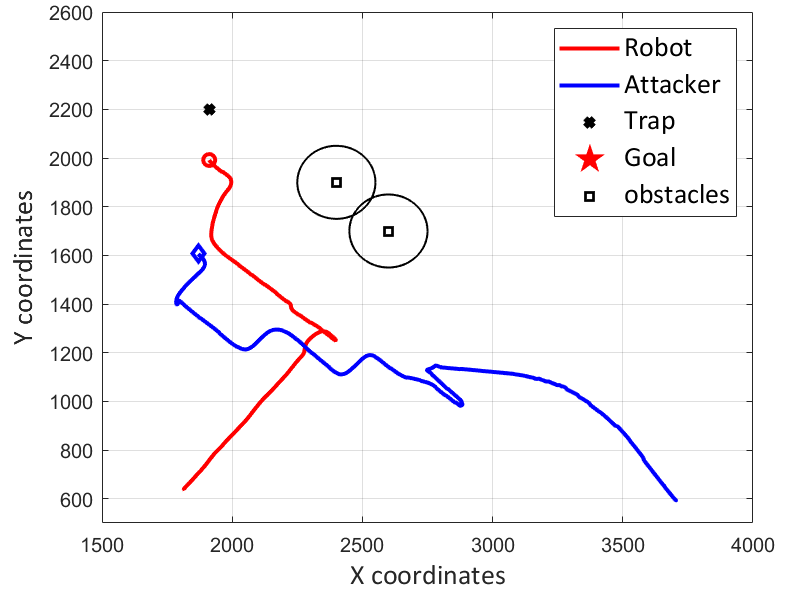}} 
\subfigure[]{\label{fig:SA_ob_speed}
\includegraphics[width=0.23\textwidth]{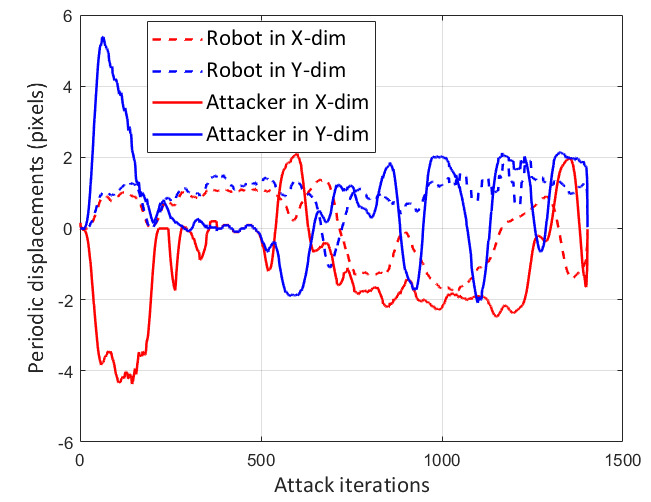}}
\vspace{-5pt}
\caption{Illustration of the shortest-path attack with other obstacles involved. 
(a) Snapshot during the attack process, where both $\text{R}_\text{a}$ and $\text{R}_\text{v}$ needs to avoid the obstacles nearby. 
(b) Snapshot when $\text{R}_\text{v}$ is successfully driven into the trap. 
(c) The trajectories of the robots, where the attack starts at step 490 and the ideal and actual attack path length are $\SI{1130}{pixels}$ and $\SI{1366}{pixels}$, respectively. 
(d) The periodic displacements of the robots in two directions. 
}
\label{fig:S_Attack}
\vspace{-5pt}
\end{figure}

\begin{figure}[t]
\subfigure[]{\label{fig:HA_ob1}
\includegraphics[width=0.23\textwidth]{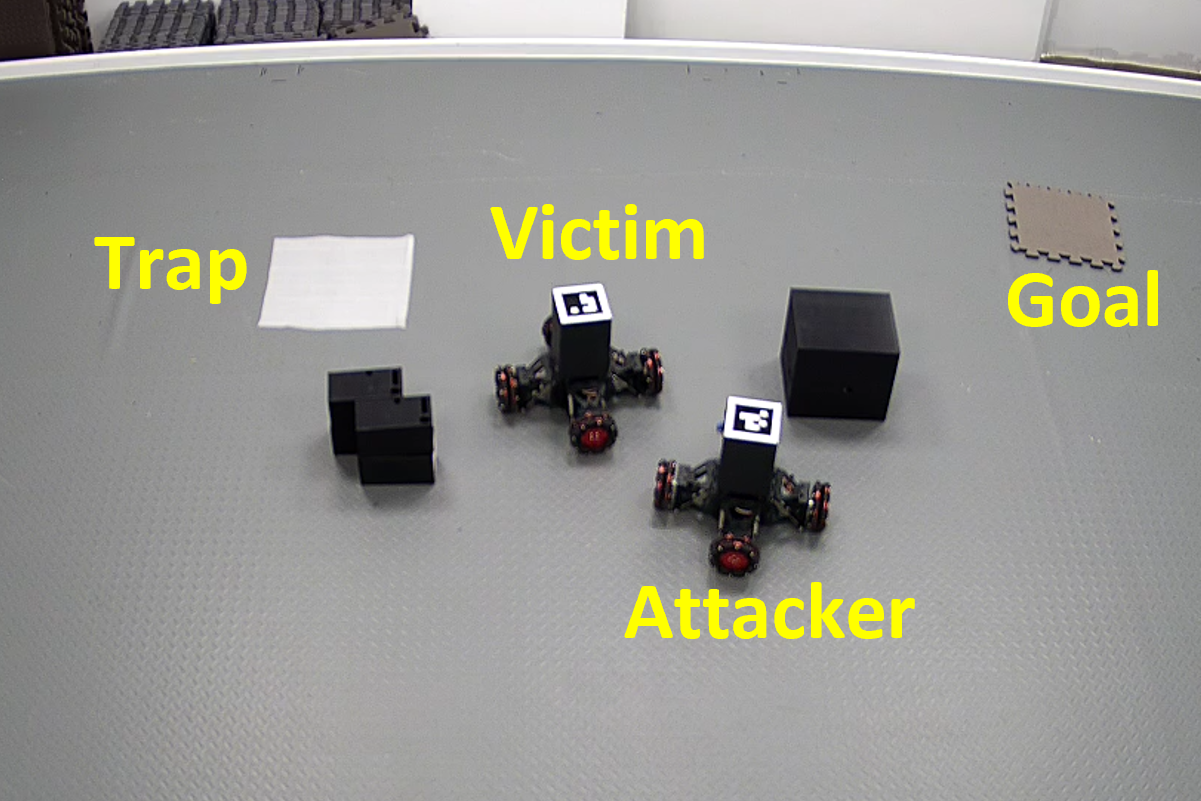}} 
\subfigure[]{\label{fig:HA_ob2}
\includegraphics[width=0.23\textwidth]{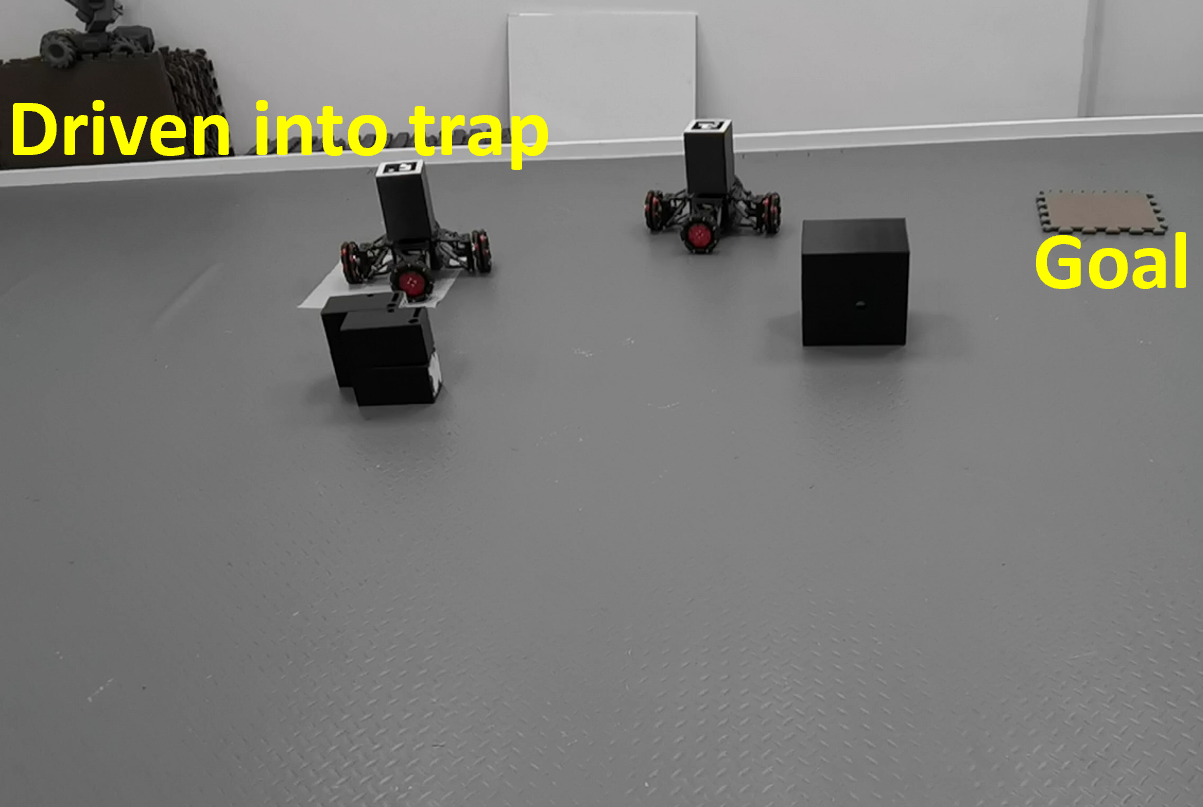}}
\subfigure[]{\label{fig:HA_ob_state}
\includegraphics[width=0.23\textwidth]{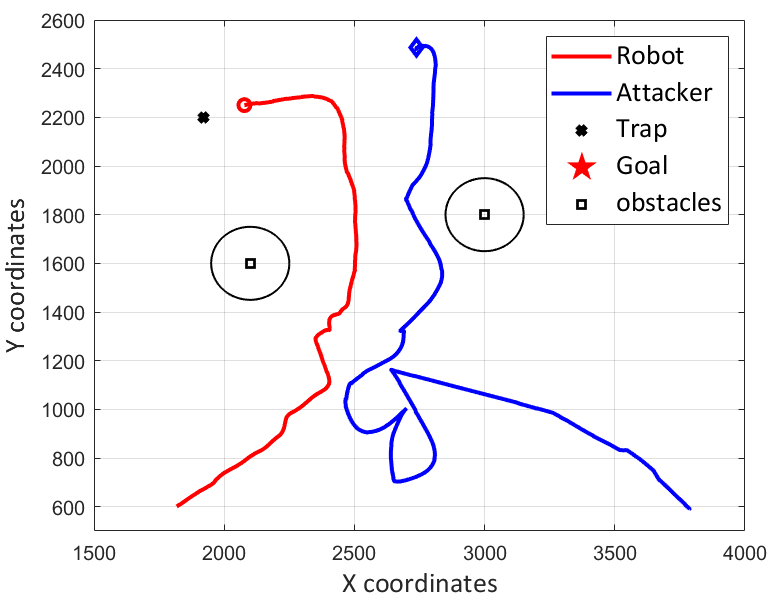}} 
\subfigure[]{\label{fig:HA_ob_speed}
\includegraphics[width=0.228\textwidth]{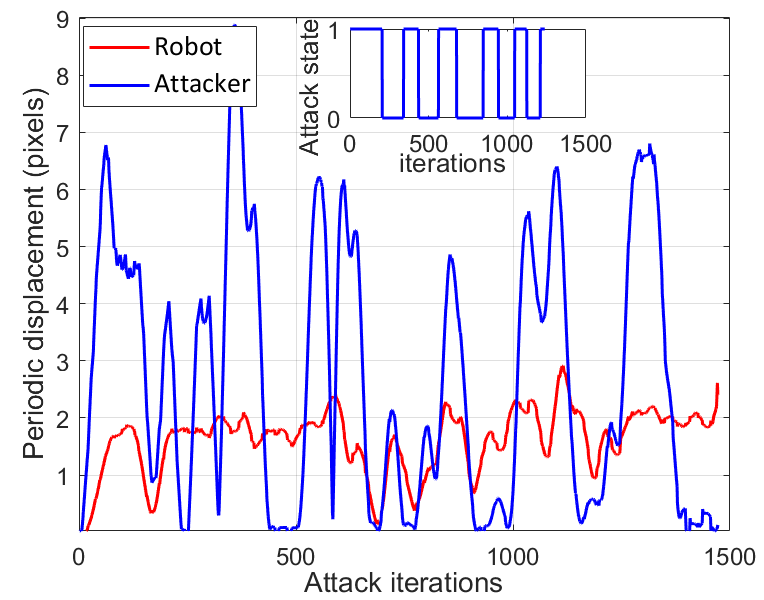}}
\vspace{-5pt}
\caption{Illustration of the hands-off attack with other obstacles involved. 
(a) Snapshot during the attack process, where both $\text{R}_\text{a}$ and $\text{R}_\text{v}$ needs to avoid the obstacles nearby. 
(b) Snapshot when $\text{R}_\text{v}$ is successfully driven into the trap. 
(c) The trajectories of the robots, where the attack starts at step 309 and the total attack horizon is $1164$ steps. 
(d) The resultant periodic displacements of the robots. The subfigure in the top right corner depicts the speed command state of $\text{R}_\text{a}$, where the active ratio of this example is $0.4356$. 
}
\vspace{-5pt}
\label{fig:H_Attack}
\end{figure}

\subsection{Experiment Results}
The first presented experiment is an illustration of the intentional learning stage, as shown in Fig.~\ref{fig:active_learning}. 
In this experiment, the attacker is assigned three observation points $[2600,1000]$, $[2900,1500]$ and $[2900,1500]$. 
Fig.~\ref{fig:learning2}-Fig.~\ref{fig:learning1} depicts the obstacle-avoidance process of the victim robot, when the attacker approaches it as an obstacle. 
The avoidance trajectory of the victim robot is observed by the attacker, and is used to construct the input-output pairs in the holonomic form of \eqref{input1}. 
Fig.~\ref{fig:learning4} shows the complete trajectory evolution during this process. 
By conducting such intentional learning procedures in different positions, we obtain 200 groups of the input-output pairs to infer the goal position by the least squares method and learn the obstacle-avoidance mechanism by the SVR method. 
The inferred obstacle-detection radius and the goal position are $\hat{D}=678~\SI{}{pixel}$ $(|\hat{D}-D|=78)$ and $\hat{p}_g=[3742,2355]$ ($\|\hat{p}_g-{p}_g\|_2=73$), while the learned model achieves mean directional accuracy of $0.91$ and root mean square error of $160.1$ pixels, which are sufficient to support the following attack implementation.

Next, we focus on the validation of the proposed shortest-path and hands-off attack strategies based on the learned model. 
To better illustrate the performance and avoid repeating similar tests multiple times, we directly provide the results considering there are also other obstacles in the environment. 
Note that we use \textit{periodic displacements} in a detection period of the cameras to describe the speed of the robots\footnote{Different from the simulation examples, the control period of the practical robots is not strictly a constant, but varies in a small range around the nominal value due to different computation burden and system noises. 
Therefore, we use the periodic displacements of the robots in two consecutive snapshots to plot the speed curve.}.

Fig.~\ref{fig:S_Attack} presents the results of the shortest-path attack, where the coordinates of the obstacles are $[2400,1900]$ and $[2600,1700]$, respectively. 
The snapshots of running process are shown in Fig.~\ref{fig:SA_ob1} and \ref{fig:SA_ob2}, while the trajectory and speed evolution of the robots are shown in Fig.~\ref{fig:SA_ob_state} and \ref{fig:SA_ob_speed}. 
During the shortest-path attack, the attacker will consistently make moves around the victim robot. 
In this example, the attack state starts at step $490$, and the ideal attack path length is $\| [1900,2200] - [2670,1373] \|_2=\SI{1130}{pixels}$, while the practical path length is $\SI{1366}{pixels}$. 
Note that this path deviation is mainly caused by the obstacles near the ideal path. 
Finally, we move to the example of the hands-off attack, where the coordinates of the obstacles are $[2400,1900]$ and $[2600,1700]$. 
Fig.~\ref{fig:H_Attack} presents the results of the shortest-path attack, where the coordinates of the obstacles are $[2100,1600]$ and $[3000,1800]$, respectively. 
The attack starts at step $309$ with a total attack horizon of $1164$ steps. 
During the attack process, the attacker is only active for $507$ steps in total, thus the active ratio of this example is $0.4356$, which corresponds to the conclusion in Theorem \ref{th-last}. 
All the experiments demonstrate the effectiveness of the proposed attack. 
More experiment tests can be found in the attached video.

\section{Conclusion}\label{sec-conclusion}
In this paper, the security problem of mobile robots is investigated. 
Different from most existing works, we propose a learning-based intelligent attack framework, where the attacker is able to learn certain dynamics of the target based on external observations, and then utilize it to accomplish its attack purpose in a more feasible and stealthier way. 
Specifically, without any prior information, the attacker is able to infer the victim robot's goal position and obstacle-detection area firstly. 
Then, the obstacle-avoidance mechanism is regressed by exploring the intrinsic characteristic of the mechanism and dynamically collecting reliable data in trials. 
Furthermore, two kinds of efficient control strategies for short path and high concealment attack objectives are designed, respectively. 
The feasibility analysis of the attacks and their performance bounds are provided, and the conditions for a successful attack implementation are proved. 
Extensive simulations and real-life experiments confirm the effectiveness of the proposed intelligent attack framework. 

There are still many open issues that require further investigations.  
For example, we mainly focus on attack design in this paper, without more detailed consideration of defense for the victim robot. 
How to constitute an integrated framework containing both attack and defense needs further investigation. 
Meanwhile, the scenario considered in this paper is relatively simple, and applying the idea to more complicated scenarios is quite meaningful, like multi-robot coordination tasks. 
More generally, questions like what other dynamics information can be learned from external observation and what sophisticated attacks can be designed are also worth further exploration.

\section*{Acknowledgement}
The authors would like to thank Xuda Ding and Zitong Wang for their help in the attack experiments.

\appendix
\subsection{Proof of Theorem \ref{th-0001}}\label{pr:th-0001}
\begin{proof}

First, we prove that the trajectory can be approximated accurately in the asymptotic sense. 
Recalling $y_{lj}=\bm{P}_l^\mathsf{T} \bm{\tilde{x}}_{lj} + \eta_{lj}$ and the trajectory $\mathcal{T}_r^{(l)}$ is a straight line, thus the coefficient vector $\bm{P}_l$ is two-dimensional. 
For simple expressions, the variables are organized as $X_l =[\bm{\tilde{x}}_{l1},\bm{\tilde{x}}_{l2},\cdots,\bm{\tilde{x}}_{l n_t}]^\mathsf{T}$, $Y_l =[y_{l1},y_{l2},\cdots,y_{l n_t}]^\mathsf{T}$ and $\Gamma_l=[\eta_{l1},\eta_{l2},\cdots,\eta_{ln_t}]^\mathsf{T}$. 
Then, the problem is transformed to to solve
\begin{equation}\label{eq:th-001-eq1}
\mathop {\min }\limits_{ P_l } ~ \| Y_l - \tilde{X}_l P_l\|_2^2 ,
\end{equation}
where its least square solution and error vector are given by 
\begin{align}
\hat{P}_l&=(\tilde{X}_l^\mathsf{T} \tilde{X}_l)^{-1}\tilde{X}_l^\mathsf{T} Y_l, \\
 E_l &=\hat{P}_l-P_l=(\tilde{X}_l^\mathsf{T} \tilde{X}_l)^{-1}\tilde{X}_l^\mathsf{T} \Gamma_l, \label{eq:p-error}
\end{align}
Note that the trajectory is limited and each element in $X_l$ is bounded, and one has $\|\tilde{X}_l^\mathsf{T} \tilde{X}_l\|_2=\bm{O}(n_t)$ and $\|(\tilde{X}_l^\mathsf{T} \tilde{X}_l)^{-1}\|_2=\bm{O}(1/n_t)$. 
Then, denote the element with largest absolute value in $X_l$ as $\tilde{x}_{\max}$, and one has 
\begin{equation}
\left\{
\begin{aligned}
&\mathbb{E}\left\{\frac{ [\tilde{X}_l^\mathsf{T} \Gamma_l]_{ij} }{n_t} \right\}=  \frac{1}{n_t}\sum\limits_{n'_t = 1}^{ n_t }  x_{n'_t}^{i} \mathbb{E}[\eta_{n'_t}^{j}] = 0, \\ 
&\mathbb{D}\left\{\frac{ [\tilde{X}_l^\mathsf{T} \Gamma_l]_{ij} }{n_t}\right\}=\sum\limits_{n'_t = 1}^{n_t } ( \frac{  [\tilde{X}_l]_{n'_ti} }{n_t})^2 \sigma^2 \le \frac{ x_{\max}^2 \sigma^2}{n_t}. 
\end{aligned}\right.
\end{equation}
By the famous Chebyshev inequality, given arbitrary $\epsilon >0$, one has 
\begin{align}\label{eq:cb}
\Pr\left\{ |\frac{ [\tilde{X}_l^\mathsf{T} \Gamma_l]_{ij} }{n_t}| < \epsilon \right\} \! \ge \! 1 \!-\! \frac{\mathbb{D}[\frac{ [\tilde{X}_l^\mathsf{T} \Gamma_l]_{ij} }{n_t}]}{\epsilon^2} \!\ge \! 1 \!-\! \frac{x_{\max}^2  \sigma^2}{n_t \epsilon^2}.
\end{align}
Let $\epsilon=(n_t)^{\frac{1}{4}}$ and substitute it into (\ref{eq:cb}), and one infers that 
\begin{align} \label{eq:pr-con}
\mathop {\lim }\limits_{n_t \to \infty } \Pr\{ \| E_l \|_2 =0 \} = 1. 
\end{align}
Since $p_g$ locates on the noise-free $\mathcal{T}_r^{(l)}$, it follows from (\ref{eq:pr-con}) that $\mathop {\lim }\limits_{{n_t} \to \infty } d_T(\mathcal{T}_r^{(l)}, {p_g}) = 0$, which proves the first statement. 

Next, consider obtaining the goal ${p}_g$ from the estimated trajectories $\{\mathcal{T}_r^{(l)}, l=1,2,\cdots, n_z \}$. 
Let $\hat{P}_l=[{\tilde a}_{l},{\tilde b}_l]^\mathsf{T}$, $\tilde{A}=[\tilde{A}_0,-\bm{1}_{n_z}]$ ($\bm{1}_{n_z}$ represents all-one vector), $\tilde{A}_0=[{{\tilde a}_1}, \cdots ,{{\tilde a}_{n_z}}]^\mathsf{T}$ and $\tilde{B} = [{{\tilde b}_1}, \cdots ,{{\tilde b}_{n_z}}]^\mathsf{T}$. 
Then, $p_g$ can be obtained by solving  
\begin{equation}\label{eq:th-001-eq2}
\mathop {\min }\limits_{ {p}_{g} } \| { \tilde{A} {p}_{g} - \tilde{B} }\|_2^2,
\end{equation}
Like (\ref{eq:th-001-eq1}), the least square solution of (\ref{eq:th-001-eq2}) is given by 
\begin{equation}
\hat{p}_g=(\tilde{A}^\mathsf{T} \tilde{A})^{-1}\tilde{A}^\mathsf{T} \tilde{B}. 
\end{equation}
Recall that the error vector ${ \tilde{A} {p}_{g} - \tilde{B} }$ is also Gaussian according to (\ref{eq:p-error}). 
Therefore, the estimation error $\|\hat{p}_g-p_g\|_2$ will goes to zero when $Z\to\infty$. 
The analysis is the same as that of $\hat{P}_l$ and the details are omitted here. 
The proof is completed. 
\end{proof}

\subsection{Proof of Theorem \ref{th:convergence}}\label{pr:convergence}
\begin{proof}
First, we prove that $p_t \!\notin\! {A}_{z1}\backslash\mathcal{A}_{z2}$ is a necessary condition by a limit performance analysis and logical deduction. 
The nature of the proposed attack to generate feasible $p_a$ whose impact could offset the impact of $p_g$ while lead $\text{R}_\text{v}$ to $p_t$. 
In the proposed algorithms, this is done by keeping $p_t$ and $\{p_a,p_g\}$ stay in different sides of $\mathcal{T}_m(p_v)$ after the attack begins, i.e., 
\begin{equation}\label{eq-defi3}
\left\{
\begin{aligned}
&I_T(\mathcal{T}_m(p_v),p_a)\!\cdot\!I_T(\mathcal{T}_m(p_v),p_g)>0,\\
&I_T(\mathcal{T}_m(p_v),p_a)\!\cdot\!I_T(\mathcal{T}_m(p_v),p_g)\!\cdot\! I_T(\mathcal{T}_m(p_v),p_t)\!<\!0. 
\end{aligned}
\right.
\end{equation}
In the extreme case where $\text{R}_\text{v}$ circumvents the obstacle with minimal curvature radius $r_{\min}$, if $p_t\in\mathcal{A}_u$, then there exists a certain moment $k_c$ when $\mathcal{T}_m(p_v)$ is parallel with $\mathcal{T}_{l1}$ (or $\mathcal{T}_{l2}$). 
After $k_c$, $p_a$ and $p_g$ are not at the same side of $\mathcal{T}_m(p_v)$, i.e., (\ref{eq-defi3}) does not hold, indicating that the impact of $p_a$ to offset the impact of $p_g$ reversely change and the movement of $\text{R}_\text{v}$ is determined by $p_g$. 
Then, it follows that $\text{R}_\text{a}$ cannot impose desired impact on $\text{R}_\text{v}$ and the proposed algorithms cannot work to achieve the attack. 
On the contrary, for $p_t\notin\mathcal{A}_u$, an attack trajectory, which satisfies(\ref{eq-defi3}) while does not meet the extreme obstacle-avoidance behavior indicated by $r_{\min}$, can always be found by Algorithm \ref{algo-3} and Algorithm \ref{algo-4}.

Next, we prove the convergence of Algorithm \ref{algo-3} from the optimization perspective when . 
Note that Algorithm \ref{algo-3} can be seen as a kind of one-dimension search method, and iteratively solves the following problem 
\begin{align} \label{eq:op}
\textbf{P}_\textbf{3}:
\mathop{\min }\limits_{ p_v }~~F(p_v)=\left\| p_t - p_v \right\|_2^2. 
\end{align}
Apparently, the global optimal solution of $\textbf{P}_\textbf{3}$ is $p_v= p_t $ and this solution is unique. 
Therefore, we need to prove that the sequence $\{p_v(k)\}$ produced by Algorithm \ref{algo-3} will converge to $p_t$ as $k$ goes to infinity. 

Note that the iterative searching process is of the form 
\begin{equation}
p_v(k+1)=p_v(k)+\lambda_k S_k,
\end{equation}
where vector $S_k$ is called the search direction, and $\lambda_k\ge0$ is called the step size. 
Similar to classic gradient-descent method, Algorithm \ref{algo-3} is essentially updating the search direction and step size. 
The feasible choice of the search direction is determined by the movement of $\text{R}_\text{a}$. 
For simple expression, we denote the search procedures of Algorithm \ref{algo-3} (i.e., Line 7-15) by a mapping $\phi$: $\mathbb{R}^2 \to \mathbb{R}$. 
Specifically, $\phi$ is equivalent to the following composite mapping, 
\begin{align}
\phi =\arg \mathop {\min }\limits_{ p_a } \{ \phi_0 (S_k) : S_k \!=\!\hat{p}_v(p_a) -p_v(k), p_a \!\in\! \mathcal{A}_a^f \}, 
\end{align}
where the inner mapping $\phi_0$ represents the step of updating $\lambda_k$, and is given by 
\begin{align}
\phi_0=\mathop {\min }\limits_{\lambda} \{ h(p_v(k)+\lambda S_k): \lambda \ge 0, \lambda\| S_k \|_2 \!\le\! \mu \}, 
\end{align}

With the feasible attack set $\mathcal{A}_a^f$ determined, $S_k=\hat{p}_v(p_a)-p_v(k)$ is a descent search direction of $F(p_v)$, i.e., 
\begin{equation}
S_k^\mathsf{T} \nabla F(p_v(k))<0. 
\end{equation}
Then, one infers that $d_{vt}(k+1)\!\le\! d_{vt}({k})$ and 
\begin{equation}
F(p_v(k+1)) \le F(p_v(k)). 
\end{equation}
Combining the differentiability and convexity of $F(p_v)$, it follows that $\phi$ is a closed mapping when $p_v\neq p_t$, and $F(p_v)$ is a descent function about $p_t$ and $\phi$. 
By the convergence theorem of search algorithms (refer to Chapter 7 in \cite{chong2004introduction}), the two conditions ensure the sequence $\{p_v(k)\}$ produced by $\phi$ will converge to $p_t$. 
The proof is completed. 
\end{proof}

\subsection{Proof of Theorem \ref{th3}}\label{pr:th3}
\begin{proof}
In essence, the motion control of a mobile robot is in discrete form, 
thus we consider formulating the problem using breadth traversal analysis. 
All the solutions of the problem are represented by a tree, where the initial attack position $p_a(a_s+1)$ is the root node and a node in floor $k'$ denotes the attack inputs at attack iteration $k'$. 

First, we prove the existence of a feasible move in each iteration. 
A feasible move means $\text{R}_\text{v}$ will move closer to $p_t$, which can be achieved by making $\text{R}_\text{a}$'s position satisfy
\begin{equation}\label{pf-th-1}
\left\{
\begin{aligned}
&{p_a}\!\in\! \mathcal{A}_d({p_v}),\\
&I_T(\mathcal{T}_m(p_v),p_a)\!\cdot\! I_T(\mathcal{T}_m(p_v),p_t)<0.
\end{aligned}
\right.
\end{equation}
Apparently, there are multiple choices of $p_a$ in each iteration such that (\ref{pf-th-1}) holds, and the next step is to determine the best $p_a$ to implement the attack. 

Based on (\ref{eq-pro-b}), multiple attack inputs from $\{u_a\!:\!{\left\| {u_a(k')} \right\|_2} \!\le\! \mu\}$ at attack step $k'$ are  sampled. 
Suppose the number of sampling groups is $n$, for every two adjacent node ${u_a^i(k')}$ and ${u_a^{i-1}(k')}$, the deviation $\left\| {u_a^i(k') - u_a^{i - 1}(k')} \right\|_2$ is the same. 
Then, we define the sub-node set of $u_a(k')$ as ${S_n}(u_a(k'))$, whose cardinal number is $n$. 
Due to the constraint (\ref{eq-pro-d}), $\text{R}_\text{a}$ can not stay unmoved all the time. 
Besides, at iteration $k'$, $u_a(k'+1)$ is computed such that the distance $\hat d(k+1)={\left\| {{\hat p_v}(k + 1) - p_t} \right\|_2} \le d(k)={\left\| {{p_v}(k) - p_t} \right\|_2}$. 
As a consequence of the two factors, 
it is deduced that the length $H$ of ${\bm{u}_{a,0:H}}$ is finite. 
Denote the maximum attack depth as ${\bar H}$ and construct a solution as 
\begin{equation}
{\bm{\tilde u}_{a,0:H}}=\left\{ {{{\tilde u}_a}(1),{{\tilde u}_a}(2), \cdots ,{{\tilde u}_a}(H)} \right\},
\end{equation}
where ${{\tilde u}_a}(k'+1)\in {S_n}(u_a(k')),~k'=1, 2,\cdots, H-1$. 
By the constraints (\ref{eq-pro-d}) and (\ref{eq-pro-e}), many sub-nodes of a node $u_a(k')$ are excluded in Algorithm \ref{algo-3}. 
Therefore, the total number of all feasible solutions $n_f\ll {n^{\bar H}}$. 

Given a sampling size $n$, let ${\bm{\tilde u}^{*,n}_{a,0:H}}$ be the best solution among $n_f$ feasible solutions, and it is intuitive to have
\begin{equation}
C_s({\bm{\tilde u}^{*,n+1}_{a,0:H}})\le C_s({\bm{\tilde u}^{*,n}_{a,0:H}}), 
\end{equation}
which is monotonically decreasing towards $C_s^*$ with $n$ growing. 
Therefore, when $n$ is large enough, there exists a ${\bm{\tilde u}^{*,n+1}_{a,0:H}}$ (not necessarily unique) such that $|C_s({\bm{\tilde u}^{*,n+1}_{a,0:H}})-C_s*|\le\varepsilon$. 
The proof is completed. 
\end{proof}

\subsection{Proof of Theorem \ref{th004}}\label{pr:th004}
\begin{proof}
We prove this result by two steps. 
First, supposing the reaction radius $r$ of $\text{R}_\text{v}$ keeps unchanged under an attack pattern of $\text{R}_\text{a}$, we establish the bounds of $l_{path}$ in the situation. 
Then, we scale the value of $r$ to obtain the final bounds. 

\textit{Step 1: Establish the bounds with unchanged $r$}. 

First, consider that $\text{R}_\text{a}$ attacks $\text{R}_\text{v}$ by being at the same relative position with it. 
This process continues until the orientation of $\text{R}_\text{v}$ is heading towards $p_t$, the path length of this process is denoted as $l_{path1}$. 
Then $\text{R}_\text{a}$ moves along with $\text{R}_\text{v}$ such that $\text{R}_\text{v}$ goes straightforward to $p_t$. 
The path length is of this process is $l_{path2}$. 
The pattern is represented by Pattern 1, shown in Fig.~\ref{pattern1}, and the angle between the second line and $\mathcal{T}_a$ is 
\begin{equation} \label{eq-phi}
\xi  = \arcsin \left( {\frac{{{r}}}{{d_{te} - {r}}}} \right).
\end{equation}
By the calculation formula for arc length, we obtain $l_{pathA1}=(\pi/2+\xi)r$, and $l_{pathA2}=(d_{te}-r)\cos\xi$. 
Then we have 
\begin{equation}
l_{pathA}=(\pi /2 + \xi  - \cos \xi )r + {d_{te}}\cos \xi.
\end{equation}

Next, consider another attack pattern such that $\text{R}_\text{v}$ moves two circular arcs, and the end is tangent to the line $\mathcal{T}_a$. 
The pattern is represented by Pattern 2, shown in Fig.~\ref{pattern2}. 
By the geometrical representation, the length of the arc part $\text{R}_\text{v}$ passed is computed as $\frac{7}{6}\pi r$, 
and the length of the straight segment part is $d_{te}-(1+\sqrt{3})r$. 
Then, we obtain
\begin{equation}
l_{pathB}=(\frac{7}{6}\pi  - 1 - \sqrt 3 )r + {d_{te}}.
\end{equation}
Note that $d_{te}>(\sqrt{3}+1)r$ and $0 < \xi  < \arcsin \frac{{\sqrt 3 }}{3}$, then 
\begin{small}
\begin{align}\label{eq-le}
{l_{pathB}} - {l_{pathA}} &= (\frac{{2\pi }}{3} - 1 - \sqrt 3  - \xi  + \cos \xi )r + {d_{te}}(1 - \cos \xi ) \nonumber \\
 &\ge (\frac{{2\pi }}{3} - \xi  - \sqrt 3 \cos \xi )r > 0.16r > 0.
\end{align}
\end{small}
\!\!(\ref{eq-le}) implicates $l_{pathB}>l_{pathA}$ always holds. 
In fact, if the reaction radius of $\text{R}_\text{v}$ is unchanged, the length of $\text{R}_\text{v}$'s path is bounded by $l_{pathA}$ and $l_{pathB}$, i.e.,
\begin{equation}\label{eq-le0}
(\pi/2+\xi  - \cos \xi )r + {d_{te}}\cos \xi  \le {l_{path}} \le (\frac{7}{6}\pi  - 1 - \sqrt 3 )r + {d_{te}}. 
\end{equation}

\textit{Step 2: Scale $r$}. 

Given a $r\in [r_{\min},r_{\max}]$, the path length is bounded by (\ref{eq-le0}). 
Since the path length is monotonically increasing with $r$ increasing, the minimum path length is obtained when $r=r_{\min}$ and the trajectory is Pattern 1. 
The maximum path length is obtained when $r=r_{\max}$ and the trajectory is Pattern 2. 
By the design of Algorithm \ref{algo-3}, even though the reaction radius of $\text{R}_\text{v}$ is time-varying, it is also bounded in $[r_{\min},r_{\max}]$. 
Then, it is induced that the real path length is between the minimum and the maximum path length. 
Due to $C_s^*=d_{te}$, we have 
$(\pi/2+\xi  \!-\! \cos \xi ){r_{\min }} + d_{te}(\cos \xi  - 1) \!\le\! \bar C_s - {C_s^*} \!\le\! (\frac{7}{6}\pi  - 1 - \sqrt 3 ){r_{\max }}$. 
The proof is completed. 
\end{proof}

\subsection{Proof of Theorem \ref{th-last}}\label{pr:th-last}
\begin{proof}
The performance bounds of Algorithm \ref{algo-4} can be obtained by resorting to analyzing a relaxed version of Algorithm \ref{algo-3}, where the $\text{R}_\text{a}$ does not have to be in the obstacle detection area all the time. 
Once the attack begins, $\text{R}_\text{a}$ does the same procedure of [Line 12-20, Algorithm \ref{algo-3}] to move, and the next iteration $\text{R}_\text{a}$ stays inactive. 
By alternately running the two operations (equivalent to one attack move in every two iterations), the attack can still be achieved with a longer attack horizon. 
Note that at every inactive state, $\text{R}_\text{a}$'s impact on $\text{R}_\text{v}$ does not perish but only work a little weaker than continuing its attack, i.e., $\text{R}_\text{v}$ will still move towards $p_t$. 
Therefore, the attack horizon $H\le2H_1$ in Algorithm \ref{algo-4}, 
with the active attack period ${\left\| {{\bm{u}_{a,0:H}}} \right\|_0}=0.5H$. 

Next, by further relaxing the attack mode such that $\text{R}_\text{v}$ always stays inactive if $\text{R}_\text{a}$'s impact still works while its movement limit is not violated, then we have ${\left\| {{\bm{u}_{a,0:H}}} \right\|_0}<0.5H$. 
The proof is completed. 
\end{proof}

\begin{IEEEbiography}[{\includegraphics[width=1in,height=1.5in,clip,keepaspectratio]{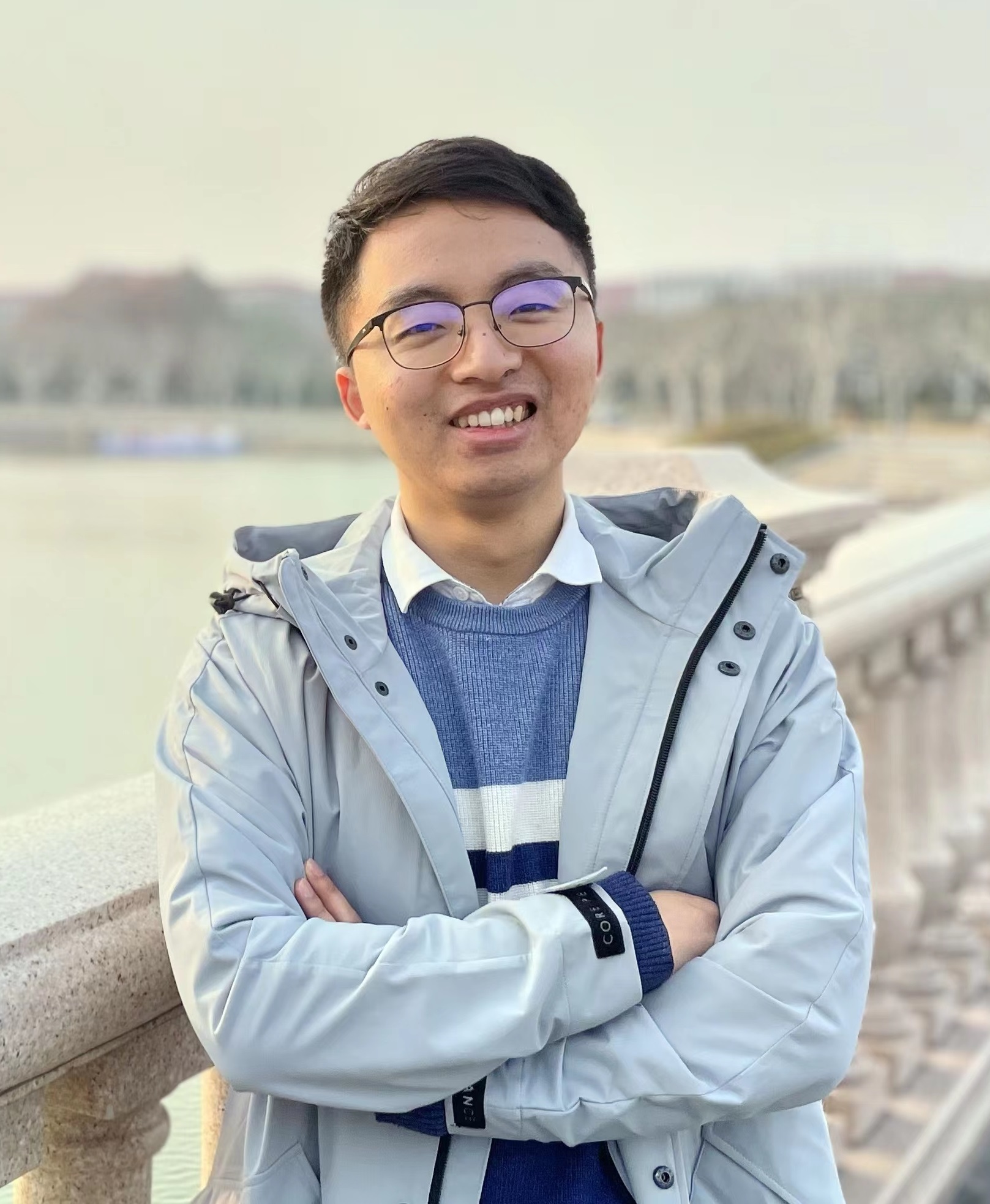}}]{Yushan Li}
(S'19) received the B.E. degree in School of Artificial Intelligence and Automation from Huazhong University of Science and Technology, Wuhan, China, in 2018. 
He is currently working toward the Ph.D. degree with the Department of Automation, Shanghai Jiaotong University, Shanghai, China. 
He is a member of Intelligent of Wireless Networking and Cooperative Control group. 
His research interests include robotics, security of cyber-physical system, and distributed computation and optimization in multi-agent networks. 
\end{IEEEbiography}

\begin{IEEEbiography}[{\includegraphics[width=1in,height=1.5in,clip,keepaspectratio]{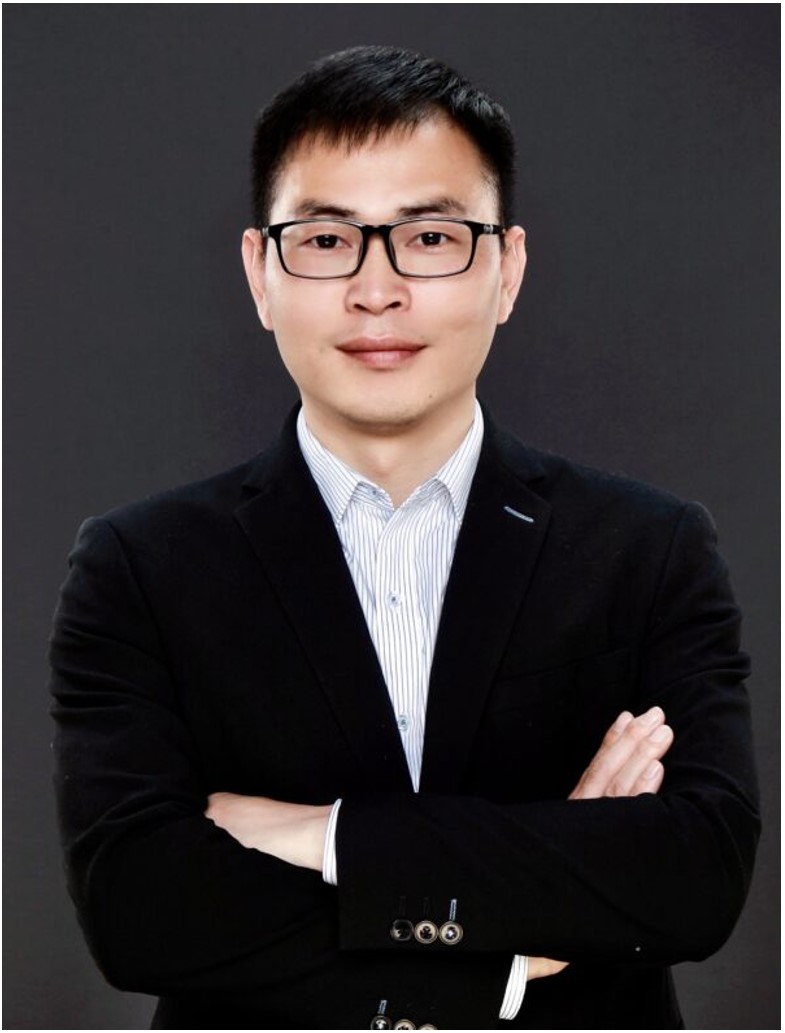}}]{Jianping He} 
(SM'19) is currently an associate professor in the Department of Automation at Shanghai Jiao Tong University. He received the Ph.D. degree in control science and engineering from Zhejiang University, Hangzhou, China, in 2013, and had been a research fellow in the Department of Electrical and Computer Engineering at University of Victoria, Canada, from Dec. 2013 to Mar. 2017. His research interests mainly include the distributed learning, control and optimization, security and privacy in network systems. 

Dr. He serves as an Associate Editor for IEEE Trans. Control of Network Systems, IEEE Open Journal of Vehicular Technology, and KSII Trans. Internet and Information Systems. He was also a Guest Editor of IEEE TAC, International Journal of Robust and Nonlinear Control, etc. He was the winner of Outstanding Thesis Award, Chinese Association of Automation, 2015. He received the best paper award from IEEE WCSP'17, the best conference paper award from IEEE PESGM'17, and was a finalist for the best student paper award from IEEE ICCA'17, and the finalist best conference paper award from IEEE VTC'20-FALL.
\end{IEEEbiography}

\begin{IEEEbiography}[{\includegraphics[width=1in,height=1.5in,clip,keepaspectratio]{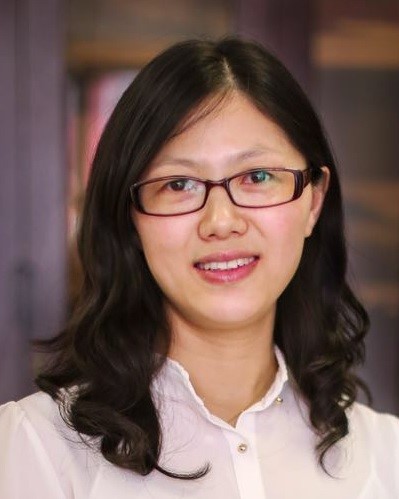}}]{Cailian Chen}
(M'06) received the B. Eng. and M.Eng. degrees in Automatic Control from Yanshan University, P. R. China in 2000 and 2002,
respectively, and the Ph.D. degree in Control and Systems from City University of Hong Kong, Hong Kong SAR in 2006. She has been with the Department of Automation, Shanghai Jiao Tong University since 2008. She is now a Distinguished Professor. 

Prof. Chen's research interests include industrial wireless networks, computational intelligence and situation awareness, Internet of Vehicles. She has authored 3 research monographs and over 100 referred international journal papers. She is the inventor of more than 20 patents. Dr. Chen received the prestigious ``IEEE Transactions on Fuzzy Systems Outstanding Paper Award'' in 2008, and ``Best Paper Award'' of WCSP'17 and YAC'18. She won the Second Prize of National Natural Science Award from the State Council of China in 2018, First Prize of Natural Science Award from The Ministry of Education of China in 2006 and 2016, respectively, and First Prize of Technological Invention of Shanghai Municipal, China in 2017. 
She was honored ``Changjiang Young Scholar'' in 2015 and ``National Outstanding Young Researcher'' by NSF of China in 2020. 

Prof. Chen has been actively involved in various professional services. She is a Distinguished Lecturer of IEEE VTS. She serves as
the Deputy Editor for National Science Open, and an Associate Editor of IEEE Transactions on Vehicular Technology, IET Cyber-Physical Systems: Theory and Applications, and Peer-to-peer Networking and Applications (Springer). She also served as the TPC Chair of ISAS'19, the Symposium TPC Co-chair of IEEE Globecom 2016, and the Track Co-chair of VTC2016-fall and VTC2020-fall.
\end{IEEEbiography}

\begin{IEEEbiography}[{\includegraphics[width=1in,height=1.5in,clip,keepaspectratio]{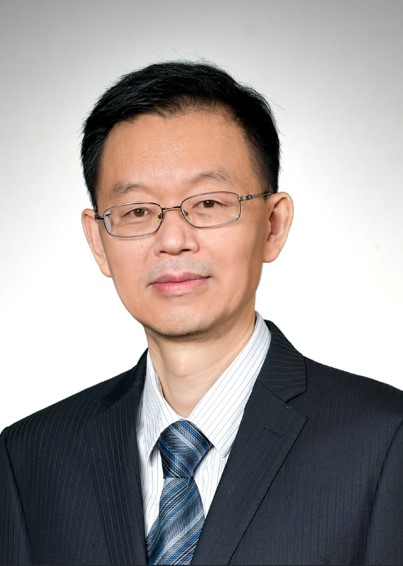}}]{Xinping Guan}
(F'18) received the B.Sc. degree in Mathematics from Harbin Normal University, Harbin, China, in 1986, and the Ph.D. degree in Control Science and Engineering from Harbin Institute of Technology, Harbin, China, in 1999. 

He is currently a Chair Professor with Shanghai Jiao Tong University, Shanghai, China, where he is the Dean of School of Electronic, Information and Electrical Engineering, and the Director of the Key Laboratory of Systems Control and Information Processing, Ministry of Education of China. Before that, he was the Professor and Dean of Electrical Engineering, Yanshan University, Qinhuangdao, China. 
Dr. Guan's current research interests include industrial cyber-physical systems, wireless networking and applications in smart factory, and underwater networks. He has authored and/or coauthored 5 research monographs, more than 270 papers in IEEE Transactions and other peer-reviewed journals, and numerous conference papers. As a Principal Investigator, he has finished/been working on many national key projects. He is the leader of the prestigious Innovative Research Team of the National Natural Science Foundation of China (NSFC). 

Dr. Guan is an Executive Committee Member of Chinese Automation Association Council and the Chinese Artificial Intelligence Association Council. He received the First Prize of Natural Science Award from the Ministry of Education of China in both 2006 and 2016, and the Second Prize of the National Natural Science Award of China in both 2008 and 2018. He was a recipient of the “IEEE Transactions on Fuzzy Systems Outstanding Paper Award” in 2008. He is a ``National Outstanding Youth'' honored by NSF of China, ``Changjiang Scholar'' by the Ministry of Education of China and ``State-level Scholar'' of ``New Century Bai Qianwan Talent Program'' of China.
\end{IEEEbiography}

\end{document}